 \documentclass[tablecaption=bottom]{jmlr}

\usepackage{booktabs}
\usepackage[normalem]{ulem}

\theorembodyfont{\upshape}
\theoremheaderfont{\scshape}
\theorempostheader{:}
\theoremsep{\newline}

\usepackage{xspace}
\usepackage{graphicx}
\usepackage{booktabs} 
\usepackage{adjustbox}
\usepackage{siunitx}
\usepackage{bm}
\usepackage[hashEnumerators,smartEllipses,hybrid]{markdown}
\usepackage{verbatim}

\newcommand{\ie}{i.e.\ }
\newcommand{\ienows}{i.e.} 
\newcommand{\eg}{e.g.\ }
\newcommand{\egnows}{e.g.} 
\newcommand{\wrt}{w.r.t.\ }
\newcommand{\iid}{i.i.d.\ }

\newcommand{\ours}{W-dropout\ }
\newcommand{\oursnows}{W-dropout} 

\newcommand{\btheta}{\theta}
\newcommand{\bthetaTilde}{\tilde{\theta}}
\newcommand{\func}{f_{\btheta}}
\newcommand{\fsub}{f_{\bthetaTilde}}

\newif\ifdraft
\usepackage{color}

\drafttrue 

\drafttrue
\ifdraft
\usepackage{cancel}
\usepackage{soul}
\newcommand{\hide}[1]{}
\fi

\jmlrproceedings{}{Arxiv Preprint}

\title[Wasserstein Dropout]{Wasserstein Dropout}

  \author{\Name{Joachim Sicking} \Email{joachim.sicking@iais.fraunhofer.de}\\
  \Name{Maram Akila} \Email{maram.akila@iais.fraunhofer.de}\\
  \Name{Maximilian Pintz} \Email{maximilian.alexander.pintz@iais.fraunhofer.de}\\
  \Name{Tim Wirtz} \Email{tim.wirtz@iais.fraunhofer.de}\\
  \addr Fraunhofer IAIS, Sankt Augustin, Germany
    \AND
  \Name{Stefan Wrobel} \Email{stefan.wrobel@cs.uni-bonn.de}\\
  \addr Fraunhofer IAIS, Sankt Augustin, Germany\\
  \addr Fraunhofer Center for Machine Learning\\
  \addr University of Bonn, Bonn, Germany
    \AND
  \Name{Asja Fischer} \Email{asja.fischer@uni-bochum.de}\\
  \addr University of Bochum, Bochum, Germany
  }

\begin{document}

\maketitle

\begin{abstract}
Despite of its importance for safe machine learning, uncertainty quantification for neural networks is far from being solved.
State-of-the-art approaches to estimate neural uncertainties are often hybrid, combining parametric models with explicit or implicit (dropout-based) ensembling. 
We take another pathway and propose a novel approach to uncertainty quantification for regression tasks, \textit{Wasserstein dropout},
that is purely non-parametric.
Technically, it captures \textit{aleatoric} uncertainty by means of dropout-based sub-network distributions.
This is accomplished by a new objective which minimizes the Wasserstein distance between the label distribution and the model distribution.
An extensive empirical analysis shows that Wasserstein dropout outperforms state-of-the-art methods, on vanilla test data as well as under distributional shift, in terms of producing more accurate and stable uncertainty estimates.

\end{abstract}

\section{Introduction}
\label{sec:intro}

Having attracted great attention in both academia and digital economy, deep neural networks (DNNs, \cite{goodfellow2016deep}) are about to become vital components of safety-critical applications. Examples are autonomous driving \citep{bojarski2016end,pomerleau1989alvinn} or medical diagnostics \citep{liu2014early}, where prediction errors potentially put humans at risk. 
These systems require methods that are robust not only under lab conditions (e.g.~\iid data sampling), but also under continuous domain shifts.
Besides shifts in the data, the data distribution itself poses further challenges.
Critical situations are (fortunately) rare and thus strongly under-represented in datasets. Despite their rareness, these critical situations have a significant impact on the safety of operations. This calls for comprehensive self-assessment capabilities of DNNs and recent uncertainty mechanisms can be seen as a step in that direction. 

\begin{figure}[bth]
    \centering
    \includegraphics[trim=0 0 0 0, clip,width=0.65\columnwidth]{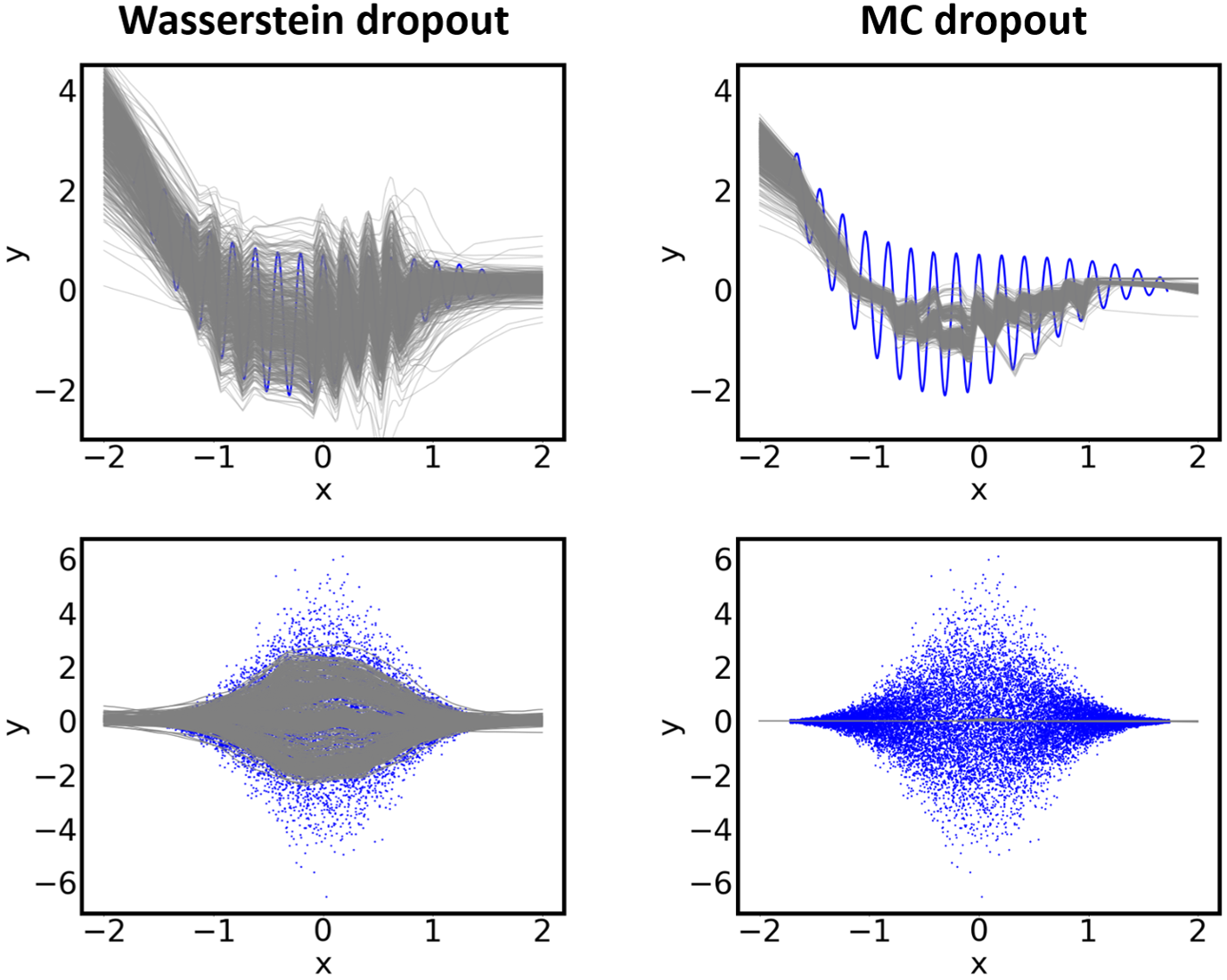} 
    \vspace*{-3mm}
    \caption{
    Wasserstein dropout (left column) employs sub-networks to model \textit{aleatoric uncertainty}, \ie
    the heterogeneous noise of (in this case, toy) datasets is reflected by the sub-network distributions of the trained models. 
    This is in contrast to other uncertainty methods like MC dropout (right column) that use sub-network distributions to model \textit{epistemic uncertainty}. 
    This type of uncertainty is small after training a model on the densely sampled toy datasets and consequently MC dropout's sub-network distributions are significantly more narrow compared to Wasserstein dropout. 
    The ground truth data is shown in blue. 
    Each gray line represents the outputs of one of 500 random sub-networks that are obtained by applying dropout-based sampling to the trained full network. For details on the data sets (`toy-hf', `toy-noise'), the neural architecture and the uncertainty methods please refer to section \ref{sec:experiments} and references therein.
    }
    \label{fig:subnets_toydata}
\end{figure}
While a variety of 
approaches to model uncertainty 
of DNN predictions in regression tasks has been established, stable  uncertainty quantification is still an open problem. 
Widely used techniques like \cite{kendall2017uncertainties} and \cite{lakshminarayanan2017simple}
combine parametric and non-parametric (ensembling-based) mechanisms 
to account for aleatoric uncertainty (data noise) and
epistemic uncertainty (model weight uncertainty).
The employed parametric mechanisms represent uncertainty estimates by dedicated network output variables, which are often interpreted as variance parameters of Gaussian distributions. These modeling techniques are sometimes also referred to as ``direct modeling'' \citep{feng2021review}.

In this work, we take a different approach and propose to model (aleatoric) uncertainty in DNNs in a novel, \textit{fully non-parametric} way. 
We introduce \textit{Wasserstein dropout (W-dropout)} that is designed to capture heteroscedastic (\ie input-dependent) data noise by means of its sub-network distribution (see Fig.\ \ref{fig:subnets_toydata}).
It builds on the idea of matching the network output distribution, resulting from randomly dropping neurons, to the (factual or implicit) data distribution by minimizing the Wasserstein distance.

In detail, we contribute
\begin{itemize}
    \item by deriving a novel and surprisingly simple Wasserstein-based learning objective for sub-networks that simultaneously optimizes task performance and uncertainty quality,
    \item by conducting an extensive empirical evaluation where \ours outperforms state-of-the-art uncertainty techniques \wrt various benchmark metrics, not only in-data but also under data shifts, 
    \item and by introducing two novel uncertainty measures: a non-saturating calibration score and a measure for distributional tails that allows to analyze worst-case scenarios \wrt uncertainty quality.
\end{itemize}

The remainder of the paper is organized as follows: first, we present related work on uncertainty estimation in neural networks in section \ref{sec:related_work}. 
Next, Wasserstein dropout is introduced in section \ref{sec:ours}. 
We study the uncertainties induced by Wasserstein dropout on various datasets in section \ref{sec:experiments}, paying special attention to safety-relevant evaluation schemes and metrics. 
An outlook in section \ref{sec:disc} concludes the paper. 

\section{Related work}\label{sec:related_work}

Approaches to estimate predictive uncertainties can be broadly categorized into three groups: Bayesian approximations, ensemble approaches and parametric models. 

Monte Carlo dropout \citep{gal2016dropout} is a prominent representative of the first group. It offers a Bayesian motivation, conceptual simplicity and scalability to application-size neural networks (NNs). This combination distinguishes MC dropout from other Bayesian neural network (BNN) approximations like in \cite{blundell2015weight} and \cite{ritter2018scalable}. A computationally more efficient version of MC dropout is one-layer or last-layer dropout (see \eg \cite{kendall2017uncertainties}). Alternatively, analytical moment propagation allows sampling-free MC-dropout inference at the price of additional approximations \mbox{(e.g. \cite{postels2019sampling})}. Further extensions of MC dropout target tuned performance by learning layer-specific drop rates using Concrete distributions \citep{gal2017concrete}, the integration of aleatoric uncertainty \citep{kendall2017uncertainties}, using a parametric approach and input-dependent dropout distributions \citep{fan2021contextual}. Note that dropout training is used\mbox{---independent} from an uncertainty context---for better model generalization \citep{srivastava2014dropout}.
An alternative sampling-based approach is SWAG which constructs a Gaussian model weight distribution from the (last segment of the) training trajectory  \citep{maddox2019simple}.

Ensembles of neural networks, so-called deep ensembles \citep{lakshminarayanan2017simple}, pose another popular approach to uncertainty modeling. Comparative studies of uncertainty mechanisms \citep{gustafsson2020evaluating,snoek2019can} highlight their advantageous uncertainty quality, making deep ensembles a state-of-the-art method.
\cite{fort2019deep} argue that ensembles capture the multi-modality of loss landscapes thus yielding potentially more diverse sets of solutions.
When used in practice, these ensembles additionally include parametric uncertainty prediction for each of their members.

The third group are the before mentioned parametric modeling approaches that extend point estimations by adding a model output that is interpreted as variance or covariance~\citep{heskes1997practical,nix1994estimating}.
Typically, these approaches optimize a (Gaussian) negative log-likelihood (NLL, \cite{nix1994estimating}) and can be easily integrated with other approaches, 
for a review see \cite{khosravi2011comprehensive}.
A more recent representative of this group is, \egnows,  
deep evidential regression \citep{amini2019deep}, which places a prior distribution on Gaussian parameters.
A closely related model class is deep kernel learning. It approaches uncertainty modeling by combining NNs and Gaussian processes (GPs) in various ways, \egnows, via an additional layer~\citep{iwata2017improving,wilson2016deep}, by using networks as GP kernels~\citep{garnelo2018neural} or by matching NN residuals with a GP~\citep{qiu2019quantifying}.

In the context of object detection, the number of applicable uncertainty methods is limited by the complexity of the employed NNs.
Nonetheless, several variants can be encountered.
For instance, MC dropout, see \eg \cite{bhattacharyya2018long} or \cite{miller2018dropout}, or parametric approaches, see \cite{he2019bounding}, can scale to network sizes relevant for such applications.
\cite{hall2020probabilistic} stress the importance of uncertainty estimation for bounding box detection.

The quality of uncertainties is typically evaluated using negative log-likelihood ~\citep{blei2006variational,gal2016dropout,walker2016uncertain}, expected calibration error (ECE,~\cite{naeini2015obtaining,snoek2019can}) and its variants and by considering correlations between uncertainty estimates and model errors, \egnows, area under the sparsification error curve (AUSE, \cite{ilg2018uncertainty}) for image tasks.
Moreover, it is common to study how useful uncertainty estimates are for solving auxiliary tasks such as out-of-distribution classification \citep{lakshminarayanan2017simple} or robustness \wrt adversarial attacks. 
An alternative approach is the investigation of qualitative uncertainty behaviors: 
\cite{kendall2017uncertainties} check if the epistemic uncertainty decreases when increasing the training set and 
\cite{wirges2019capturing} study how the level of uncertainty depends on the distance of the object to a car for some 3D environment regression task.


\section{Wasserstein dropout}\label{sec:ours}
Before we lay out our dropout-based approach to modeling \textit{aleatoric} uncertainty, we analyze some central properties of Monte Carlo dropout.
The latter also employs sub-networks, however, for the purpose of modeling \textit{epistemic} uncertainty \citep{gal2016dropout}:
Given a  neural network \mbox{$\func : \mathbb{R}^d\to \mathbb{R}^m$} with parameters $\btheta$, MC dropout samples sub-networks $\fsub$ by randomly dropping nodes from the main model $\func$ yielding for each input $x_i$ a 
distribution $\mathcal{D}_{\bthetaTilde}(x_i)$ over network predictions.
During MC dropout inference the final prediction is given by the mean of a 
sample from $\mathcal{D}_{\bthetaTilde}(x_i)$,
while the uncertainty associated with this prediction can be estimated 
as a sum of its variance and a constant uncertainty offset.
The value of the latter term requires dataset-specific optimization.
During MC dropout training, minimizing the 
objective function, \egnows, 
the mean squared error (MSE), shifts all sub-network predictions 
towards the same training targets.
For a more formal explanation of this behavior, and without loss of generality, let $\func$ be a NN with one-dimensional output. 
The expected MSE  for a training sample $(x_i,y_i)$ under the model's output distribution $\mathcal{D}_{\bthetaTilde}(x_i)$
is given by
\begin{equation}\label{eqn:mc_dropout}
E_{\bthetaTilde}\left[(\fsub(x_i) - y_i)^2
\right] = \left(
\mu_{\bthetaTilde}(x_i)
- y_i\right)^2 + \sigma_{\bthetaTilde}^2(x_i)\ ,
\end{equation}
with sub-network mean $\mu_{\bthetaTilde}(x_i) = E_{\bthetaTilde}[\fsub(x_i)]$ and variance $\sigma_{\bthetaTilde}^2(x_i) =  E_{\bthetaTilde}[\fsub^2(x_i)] \allowbreak - E_{\bthetaTilde}[ \fsub(x_i) ]^2$.
Therefore, training simultaneously minimizes the squared error between sub-network mean 
$\mu_{\bthetaTilde}(x_i)$
and target $y_i$ as well as the variance 
$\sigma^2_{\bthetaTilde}(x_i)$.

As we, in contrast, seek to employ sub-networks to model \textit{aleatoric} uncertainty, minimizing the variance over the sub-networks is not desirable for our purpose. 
Instead, we aim at 
explicitly fitting the sub-network variance $\sigma^2_{\bthetaTilde}(x_i)$ to the input-dependent, \ie heteroscedastic, data variance.
That is to say, we not only match the mean values as in (\ref{eqn:mc_dropout}) but seek to match the \textit{entire} data distribution $\mathcal{D}_y(x_i)$ by means of the model's output distribution $\mathcal{D}_{\bthetaTilde}(x_i)$. This output distribution is induced by applying Bernoulli dropout to all activations of the network. 
To measure the distance between the two distributions $\mathcal{D}_{\bthetaTilde}(x_i)$ and $\mathcal{D}_y(x_i)$, a squared 2-Wasserstein metric \citep{villani2008optimal} is employed.
As it is `transport'-based it can provide a training signal also for non-overlapping distributions\footnote{A property not given for, \egnows, the Kullback-Leibler divergence, see \cite{arjovsky2017wasserstein}.} and reduces to the ``original'' MSE loss for point masses, \ienows, in the absence of aleatoric uncertainty.
Assuming that both distributions $\mathcal{D}_{\bthetaTilde}(x_i)$ and $\mathcal{D}_y(x_i)$ are Gaussian\footnote{An assumption shared by, \egnows, the NLL optimization or the ECE. While different distributions, for example exponentially decaying or mixtures, could be used in principle, we restrict the scope here to this standard Gaussian case.} yields a compact analytical expression
\begin{align}\label{eqn:ws_dist}
\text{WS}_2^2(x_i) &=\mathrm{WS}^2_2\left[\mathcal{D}_{\bthetaTilde}(x_i), \mathcal{D}_y(x_i)\right] \nonumber\\
&=\mathrm{WS}^2_2\left[\mathcal{N}(\mu_{\bthetaTilde}(x_i), \sigma_{\bthetaTilde}(x_i)), \mathcal{N}(\mu_y(x_i), \sigma_y(x_i))\right] \nonumber\\ &=\left(\mu_{\bthetaTilde}(x_i) - \mu_{y}(x_i)\right)^2+\left(\sigma_{\bthetaTilde}(x_i) - \sigma_{y}(x_i)\right)^2 \,,
\end{align}
with $\mu_{\bthetaTilde}(x_i) = E_{\bthetaTilde}[\fsub(x_i)]$ and $\sigma_{\bthetaTilde}^2(x_i) = E_{\bthetaTilde}[(\fsub (x_i) - E_{\bthetaTilde}[\fsub(x_i)])^2]$, 
and $\mu_y,\sigma_y$ defined analogously \wrt the data distribution.

In practice however, (\ref{eqn:ws_dist}) cannot be readily used as the distribution of \(y\) given $x_i$ is typically not accessible.
Instead, for a given, fixed value of \(x_i\) from the training set only a single value of \(y_i\) is known. 
Therefore, we take $y_i$ as a (rough) one-sample approximation of the mean $\mu_y(x_i)$ resulting in $\mu_y(x_i) \approx y_i$ and \mbox{$\sigma_y^2(x_i) \approx E_y[(y - y_i)^2]$}.
However, $\sigma_y^2(x_i)$ cannot be inferred from a single sample.
Inspired by parametric bootstrapping \citep{dekking2005modern,hastie2009elements}, we therefore 
approximate
the empirical data variance (for a given mean value $y_i$ and input $x_i$) with samples from our model, \ienows,  
we approximate\\
\mbox{$E_y[(y - y_i)^2]$} by
\begin{align}
E_{\bthetaTilde}[(\fsub(x_i) - y_i)^2] 
= (\mu_{\bthetaTilde}(x_i)-y_i)^2 + \sigma_{\bthetaTilde}^2(x_i)\enspace.
\end{align}
Inserting our approximations $\mu_y(x_i) \approx y_i$ and $\sigma_y(x_i) \approx (\mu_{\bthetaTilde}(x_i)-y_i)^2 + \sigma_{\bthetaTilde}^2(x_i)$ into (\ref{eqn:ws_dist}) yields the \textit{Wasserstein dropout loss} (W-dropout) 
for a data point $(x_i,y_i)$ from the training distribution:
\begin{align}\label{eqn:exact_ws_loss}
    \text{WS}_2^2(x_i) \approx&\ (\mu_{\bthetaTilde}(x_i)-y_i)^2 \\
    &+ \left[ \sqrt{\sigma_{\bthetaTilde}^2(x_i)} - \sqrt{(\mu_{\bthetaTilde}(x_i)-y_i)^2 + \sigma_{\bthetaTilde}^2(x_i)} \right]^2\,. \nonumber
\end{align}
Considering a mini-batch of size  $M$ instead of a single data point, we arrive at the optimization objective $\text{WS}_{\textup{batch}}^2 = \frac{1}{M}\sum_{i=1}^M\text{WS}_2^2(x_i)$.
In practice, $\mu_{\bthetaTilde}(x_i) \approx \frac{1}{L}\sum_{l=1}^L f_{\tilde{\theta}_l}(x_i)$ and $\sigma_{\bthetaTilde}^2(x_i) \approx \frac{1}{L}\sum_{l=1}^L f_{\tilde{\theta}_l}^2(x_i) - (\frac{1}{L}\sum_{l=1}^L f_{\tilde{\theta}_l}(x_i))^2$ are approximated by empirical estimators using a sample \mbox{size $L$}.
In contrast to MC dropout we require thereby $L$ stochastic forward passes per data point during training (instead of one), while at inference procedures are exactly the same.

Besides the regression tasks considered here our approach could be useful for other objectives which use or benefit from an underlying distribution, \egnows, Dirichlet distributions to quantify uncertainty in classification, as discussed in the conclusion.


\section{Experiments}
\label{sec:experiments}
We first outline the scope of our empirical study in subsection \ref{sec:benchmarks_measures} and begin with experiments on illustrative and visualizable toy datasets in subsection \ref{sec:toydataset}. 
Next, we benchmark \ours on various 1D datasets (mostly from the UCI machine learning repository \citep{Dua:2019}) in subsection \ref{sec:uci}, considering both in-data and distribution-shift scenarios.
In subsection \ref{sec:squeezeDet}, \ours is applied to the complex task of object detection
using the compact SqueezeDet architecture \citep{wu2017squeezedet}.


\subsection{Benchmark approaches and evaluation measures}
\label{sec:benchmarks_measures}
In this subsection, we present the considered benchmark approaches (first paragraph) and evaluation measures for uncertainty modeling. 
Aside established measures (second paragraph), we propose two novel uncertainty scores: 
an unbounded calibration measure and an uncertainty tail measure for the analysis of worst-case scenarios \wrt uncertainty quality 
(third and forth paragraph). 
A brief overview of the technical setup (last paragraph) concludes the subsection.


\paragraph{Benchmark approaches} We compare \ours networks to archetypes of uncertainty modeling, namely approximate Bayesian techniques, parametric uncertainty, and ensembling approaches.
From the first group, we pick MC dropout (abbreviated as \textbf{MC}, \cite{gal2016dropout}) and Concrete dropout (\textbf{CON-MC}, \cite{gal2017concrete}).
The variance of MC is given as the sample variance plus a dataset-specific regularization term. The networks employing these methods do not exhibit parametric uncertainty outputs (see below). 
We additionally consider SWA-Gauss\-ian (\textbf{SWAG}, \cite{maddox2019simple}), which samples from a Gaussian model weight distribution that is constructed based on model parameter configurations along 
the (final segment of the) training trajectory.
While these sampling-based approaches integrate uncertainty estimation into the structure of the entire network, \textit{parametric} approaches model the variance directly as the output of the neural network \citep{nix1994estimating}. Such networks typically output mean and variance of a Gaussian distribution $(\mu, \sigma^2)$ 
and are trained by likelihood maximization.
This approach is denoted as \textbf{PU} for parametric uncertainty. Ensembles of PU-networks \citep{lakshminarayanan2017simple}, referred to as deep ensembles, pose a widely used state-of-the-art method for uncertainty estimation \citep{snoek2019can}. Deep evidential regression (\textbf{PU-EV}, \cite{amini2019deep}) extends this parametric approach and considers prior distributions over $\mu$ and $\sigma$.
\cite{kendall2017uncertainties} consider drawing multiple dropout samples from a parametric uncertainty model and aggregating multiple predictions for $\mu$ and $\sigma$. We denote this approach \textbf{PU-MC}.
Moreover, we consider ensembles of non-parametric standard networks. We refer to the latter ones as \textbf{DEs} while we call those using additionally PU-based uncertainty \textbf{PU-DEs}. All considered types of networks provide estimates $(\mu_i,\sigma_i)$ where $\sigma_i$ is obtained either as direct network output (PU, PU-EV), by sampling (MC, CON-MC, SWAG, \oursnows) or as an ensemble aggregate (DE, PU-DE). For PU-MC, a combination of parametric output and sampling is employed. Throughout this section, we subsume PU, PU-EV, PU-DE and PU-MC as ``parametric methods". 


\paragraph{Standard evaluation measures} In all experiments we evaluate both regression performance and uncertainty quality. 
Regression performance is quantified by the root-mean-square error $\sqrt{(1/N\,\sum_i (\mu_i-y_i)^2 }$ (\textbf{RMSE}, \cite{bishop2006pattern}). 
Another established metric in the uncertainty community is the (Gaussian) negative log-likelihood (\textbf{NLL}), $1/N \sum_i \big( \log\sigma_i$ $+ (\mu_i - y_i)^2/(2 \sigma_i^2) + c \big)$, a hybrid between performance and uncertainty measure \citep{gneiting2007strictly}, see appendix \ref{appendix:nll} for a discussion. 
Throughout the paper, we ignore the constant $c=\log\sqrt{2\pi}$ of the NLL. 
The expected calibration error (\textbf{ECE}, \cite{Kuleshov2018}) in contrast is not biased towards well-performing models and in that sense a pure uncertainty measure. 
It reads ECE $= \sum_{j=1}^B \vert \tilde{p}_j - 1/B\vert $ for B equally spaced bins in 
quantile space and $\tilde{p}_j = \vert \{r_i \vert  q_j \leq \tilde{q}(r_i) < q_{j+1}\}\vert /N$ the empirical frequency of data points falling into such a bin. 
Their \textit{normalized prediction residuals} $r_i$ are defined as $r_i = (\mu_i - y_i)/\sigma_i$. 
Further, $\tilde{q}$ is the cdf of the standard normal distribution $\mathcal{N}(0,1)$ and $[q_j,q_{j+1})$ are equally spaced intervals on $[0,1]$, \ienows, $q_j=(j-1)/B$.

\begin{figure}[th]
    \centering
    \includegraphics[width=0.80\textwidth]{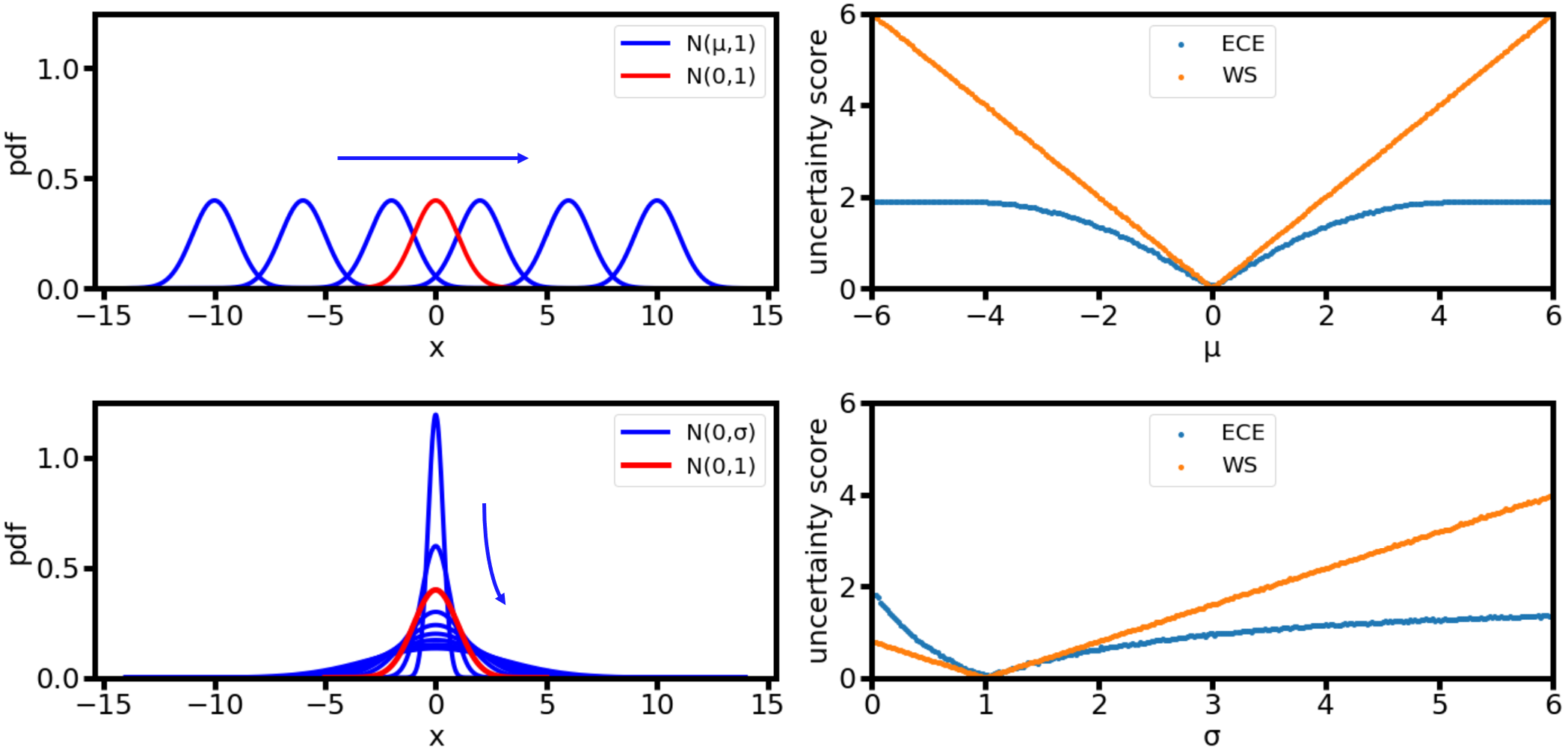} 
    \caption{Comparison of the proposed Wasserstein-based measure (WS) and the expected calibration error (ECE). We measure the deviation between a standard normal distribution $\mathcal{N}(0,1)$ (lhs, red) and shifted normal distributions $\mathcal{N}(\mu, 1)$ (top left, dark blue) and squeezed/stretched normal distributions $\mathcal{N}(0, \sigma)$ (bottom left, dark blue), respectively. The resulting ECE values (orange) and WS values (blue) on the rhs emphasize the higher sensitivity of WS in case of large distributional differences. For details on ECE and WS, see text.}
    \label{fig:ece_ws_toy_data}
\end{figure}
\begin{figure}[th]
    \centering
    \includegraphics[width=0.8\textwidth]{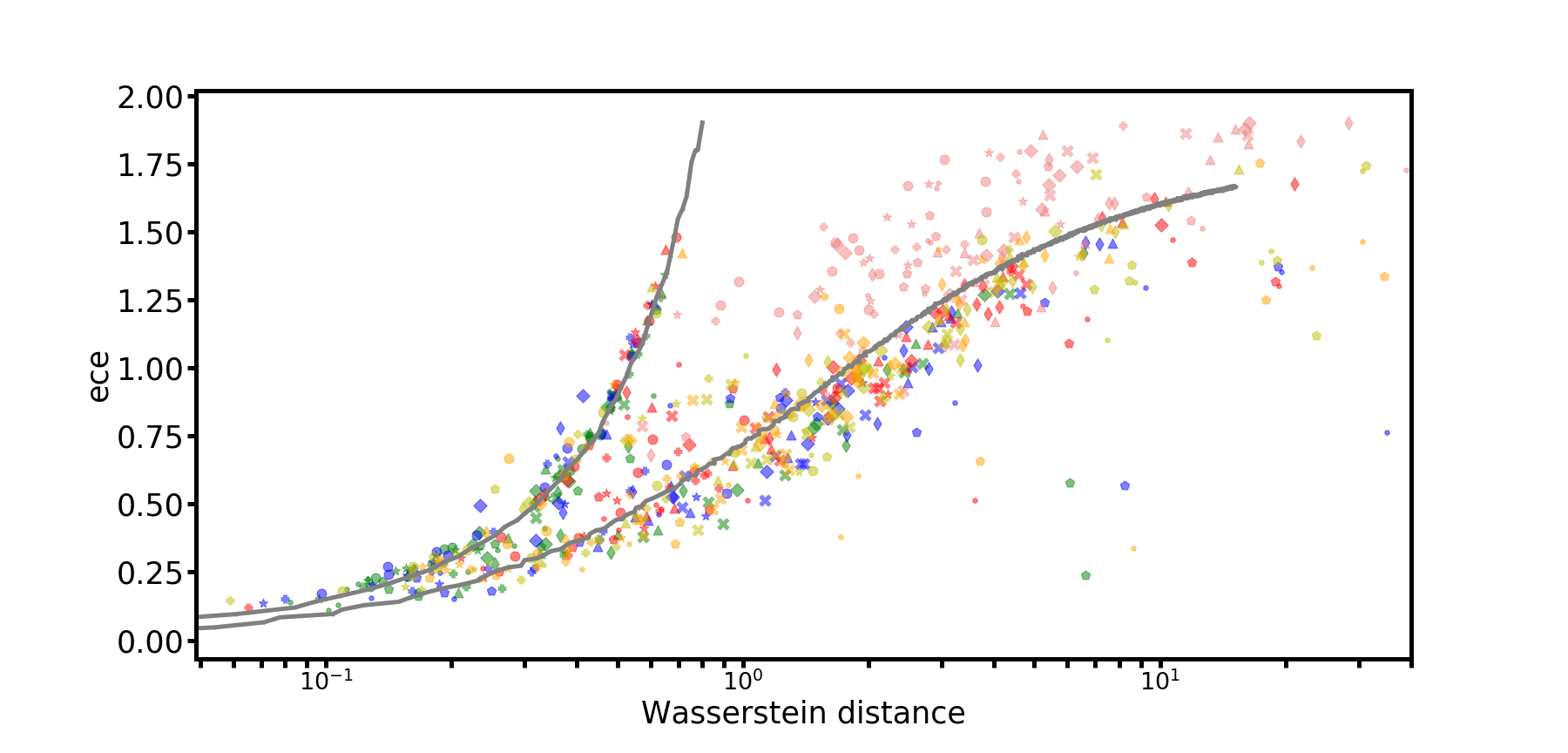} 
    \caption{Dependency between the Wasserstein-based measure and the expected calibration error for Gaussian toy data (gray curves) and for 1D standard datasets (point cloud, see subsection \ref{sec:uci} for details). The toy curves are obtained by plotting (WS$(\sigma)$, ECE$(\sigma)$) from Fig.\ \ref{fig:ece_ws_toy_data} (bottom right). For 1D standard datasets, uncertainty methods are encoded via plot markers, data splits via color. Datasets are not encoded and cannot be distinguished (see appendix \ref{appendix:unc_measures} for more details). Each plot point corresponds to a cross-validated trained network.}
    \label{fig:ece_ws_uci_data}
\end{figure}

\paragraph{An unbounded uncertainty calibration measure}
A desirable property for uncertainty measures is a signal that grows (preferentially linearly) with 
the misalignment between predicted and ideal uncertainty estimates, especially 
when handling strongly deviating uncertainty estimates.
As the Wasserstein metric fulfils this property, we not only use it for model optimization but 
propose to consider the \textit{\mbox{1-Wasserstein} distance of normalized prediction residuals} (\textbf{WS}) as a complementary uncertainty evaluation measure.
It is generally applicable and by no means restricted to \ours networks. In detail, the 1-Wasserstein distance \citep{villani2008optimal}, also known as earth mover's distance \citep{earthmover}, is a transport-based measure, denoted by $d_{\rm WS}$, between two probability densities, with Wasserstein GANs \citep{arjovsky2017wasserstein} as its most prominent application in machine learning. 
In the context of uncertainty estimation, we use the Wasserstein distance to measure deviations of uncertainty estimates $\{r_i\}_i$ from ideal (Gaussian)\footnote{As stated before, Gaussianity, while not always given, is a standard assumption for uncertainty modeling and typically used in ECE and NLL.} calibration that is given if $y_i \sim \mathcal{N}(\mu_i,\sigma_i)$ with accompanying normalized residuals of $r_i \sim \mathcal{N}(0,1)$, \ie we calculate $d_{\rm WS}\left(\{r_i\}_i,\mathcal{N}(0,1)\right)$.
As ECE, this is a pure uncertainty measure.
However, it is not based on quantiles but directly on normalized residuals and can therefore resolve deviations on all scales.
For example, two strongly ill-calibrated uncertainties 
would result in (almost) identical ECE values 
while WS would resolve this difference in magnitude. 
Let us compare ECE and WS more systematically: we consider normal distributions $\mathcal{N}(\mu, 1)$ and $\mathcal{N}(0, \sigma)$ (see Fig.\ \ref{fig:ece_ws_toy_data}) that are shifted (top left panel, dark blue) and squeezed/stretched (bottom left panel, dark blue), respectively. Their deviations from the ideal normalized residual distribution (the standard normal, red) are measured in terms of both ECE (r.h.s., blue) and WS (r.h.s., orange). For large values of $\vert \mu\vert $ and $\sigma$, ECE is bounded while WS increases linearly showing the better sensitivity of the latter towards strong deviations. For small values, $\sigma \rightarrow 0$, ECE takes its maximum value, WS a value of 1. In Fig.\ \ref{fig:ece_ws_uci_data}, we visualize these value pairs (WS$(\sigma)$, ECE$(\sigma)$) (gray lines), \ie $\sigma$ serves as curve parameter. The upper `branch' corresponds to $0<\sigma<1$, the lower `branch' to $\sigma > 1$. For comparison, the pairs (WS, ECE) of various networks trained on standard regression datasets are visualized (see subsection \ref{sec:uci} for experimental details and results). They approximately follow the theoretical $\sigma$-curve, emphasizing that both under- and overestimating variance is of practical relevance. A given WS value allows, due to lacking saturation for underestimation, to distinguish these two cases more easily compared to ECE. While one might rightfully argue that the higher sensitivity of WS leads to a certain susceptibility to potential outliers, this can be addressed by regularizing the normalized residuals or by filtering extreme outliers.



\paragraph{A novel uncertainty tail measure}
We furthermore introduce a measure for distributional tails that allows to analyze worst-case scenarios \wrt uncertainty quality, thus reflecting safety considerations.
Such potentially critical worst-case scenarios are signified by the above mentioned outliers, where the locally predicted uncertainty strongly underestimates the actual model error.
A better understanding of uncertainty estimates in these scenarios might allow to determine lower bounds on operation quality of safety-critical systems. For this, we consider normalized residuals $r_i=(\mu_i-y_i)/\sigma_i$ based on the prediction estimates $(\mu_i,\sigma_i)$ for a \mbox{given data point $(x_i,y_i)$}. 
As stated, we restrict our analysis to uncertainty estimates that \textit{underestimate} model errors, \ienows, $\vert r_i\vert \gg 1$. These cases might be more harmful than overly large uncertainties, \mbox{$\vert r_i\vert \ll 1$}, that likely trigger a conservative system behavior. We quantify uncertainty quality for worst-case scenarios as follows: for a given (test) dataset, the absolute normalized residuals $\{\vert r_i\vert \}_i$ are calculated. We determine the $99\%$ quantile $q_{0.99}$ of this set and calculate the mean value over all $\vert r_i\vert  > q_{0.99}$, the so-called \textit{expected tail loss at quantile $99\%$} ($\boldsymbol{{\rm ETL}_{0.99}}$, \cite{rockafellar2002conditional}). 
The ETL$_{0.99}$ thus measures the average uncertainty quality of the worst $1\%$.


\paragraph{Technical setup}
For the first two parts we use almost identical \mbox{setups} of 2 hidden layers with ReLu activations, using 50 neurons per layer for the toy datasets and 100 for the 1D standard datasets.
All dropout-based networks (MC, CON-MC, \oursnows) apply Bernoulli dropout to all hidden activations.
For \ours networks, we sample $L = 5$ sub-networks in each optimization step, other values of $L$ are considered in appendix \ref{appendix:hyper}. On the smaller toy datasets, we afford $L=10$.
For MC and \oursnows, the drop rate is set to $p = 0.1$ (see appendix \ref{appendix:hyper} for other values of $p$). The drop rate of CON-MC in contrast is learned during training and (mostly) takes values between $p=0.2$ and $p=0.5$. 
For ensemble methods (DE, PU-DE) we employ $5$ networks. 
All NNs are optimized using the Adam optimizer \citep{kingma2014adam} with a learning rate of $0.001$.
Additionally, we apply standard normalization to the input and output features of all datasets to enable better comparability.
The number of training epochs and cross validation runs depends on the dataset size. Further technical details on the networks, the training procedure, and the implementation of the uncertainty methods can be found in appendix \ref{appendix:experimentalSetup}.
In using a least squares regression, we make the standard assumption that errors follow a Gaussian distribution.
This is reflected in the (standard) definitions of above named measures, \ienows, all uncertainty measures quantify the set of outputs $\{(\mu_i, \sigma_i)\}$ relative to a Gaussian distribution.

\subsection{Toy datasets}
\label{sec:toydataset}

\begin{figure*}[t]
    \centering
    \includegraphics[width=\textwidth]{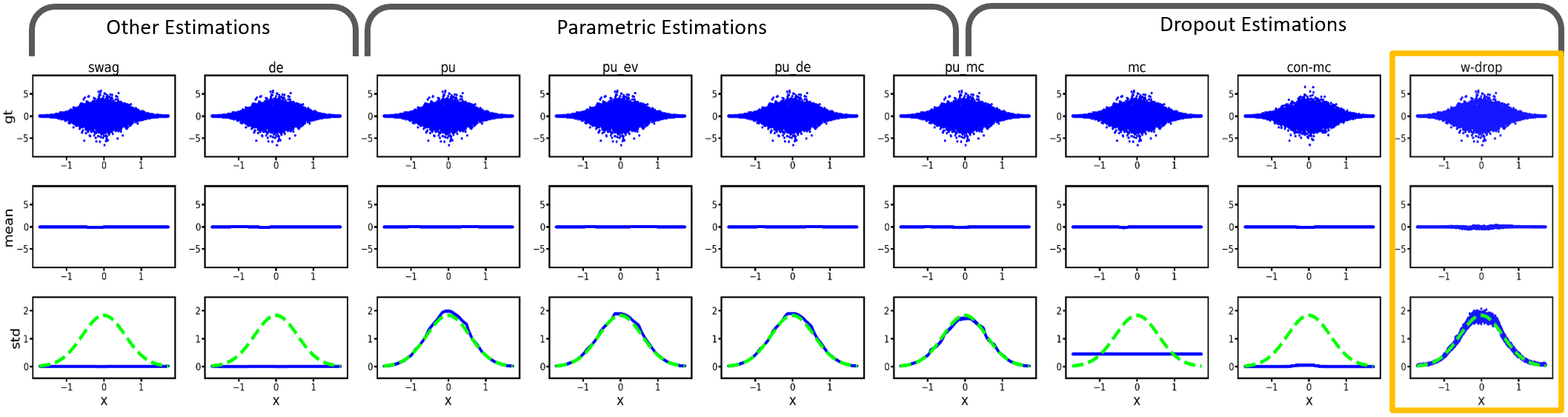} 
    \includegraphics[width=\textwidth]{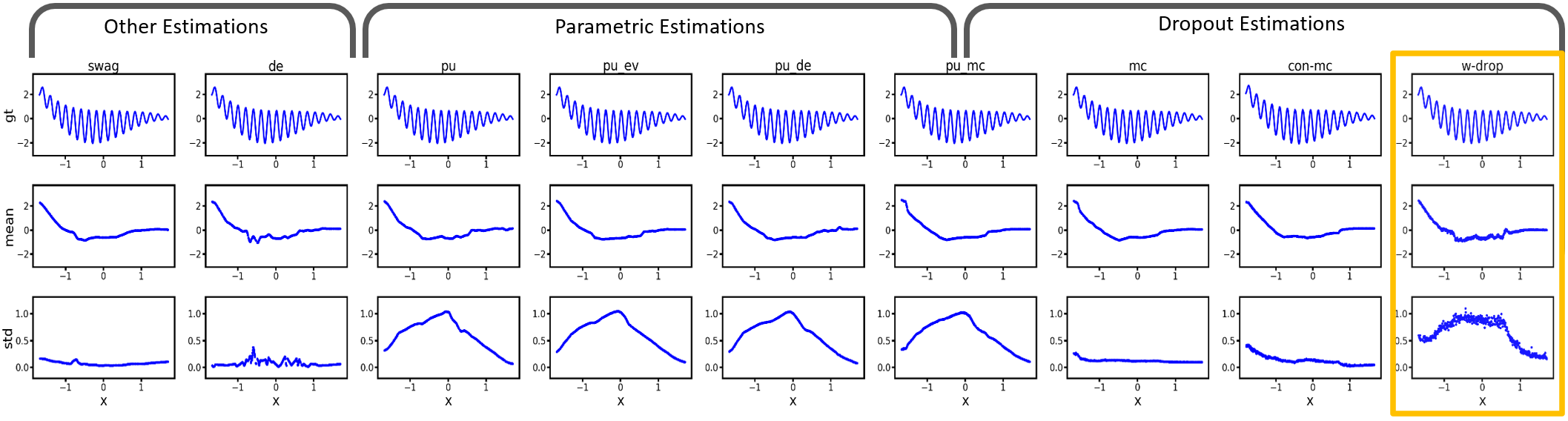} 
    \caption{
    Comparison of uncertainty approaches (columns) on two 1D toy datasets: a noisy one (top) and a high-frequency one (bottom). Test data ground truth (respective first row) is shown with mean estimates (resp. second row) and standard deviations (resp. third row). The light green dashed curve (third row) indicates the ground truth uncertainty. Similar uncertainty approaches (columns) are grouped together, \ours is highlighted by a yellow frame.}
    \label{fig:toy_hf}
\end{figure*}
To illustrate qualitative behaviors of the different uncertainty techniques, we consider two $\mathbb{R}\to\mathbb{R} $ toy datasets. This benchmark puts a special focus on the handling of aleatoric heteroscedastic uncertainty. The first dataset is Gaussian white noise with an $x$-dependent amplitude, see first row of Fig.\ \ref{fig:toy_hf}. The second dataset is a polynomial overlayed with a high-frequency, amplitude-modulated sine, see fourth row of Fig.\ \ref{fig:toy_hf}. The explicit equations for the toy datasets used here can be found in appendix \ref{appendix:toyEval}.

While the uncertainty in the first dataset (`toy-noise') is clearly visible, it is less obvious for the fully deterministic second dataset (`toy-hf').
There is an effective uncertainty due to the insufficient expressivity of the model though, as the shallow networks employed are empirically not able to fit (all) fluctuations of `toy-hf' (see fifth row of Fig.\ \ref{fig:toy_hf}). One might (rightfully) argue that this is a sign of insufficient model capacity. But in more realistic, \egnows, higher dimensional and sparser datasets the distinction between true noise and complex information becomes exceedingly difficult to make and regularization is actively used to suppress the modeling of (ideally) undesired fluctuations.
As the Nyquist-Shannon sampling theorem states,
with limited data deterministic fluctuations above a cut-off frequency can no longer be resolved \citep{landau1967sampling}. They therefore become virtually indistinguishable from random noise.

The mean estimates of all uncertainty methods (second and fifth row in Fig.\ \ref{fig:toy_hf}) look alike on both datasets. They approximate the noise mean and the polynomial, respectively. In the latter case, all methods rudimentarily fit some individual fluctuations. The variance estimation (third and sixth row in Fig.\ \ref{fig:toy_hf}) in contrast reveals significant differences between the methods:
MC dropout variants and other non-parametric ensembles are not capable of capturing heteroscedastic aleatoric uncertainty.
This behavior of MC is expectable as it was primarily introduced to account for model uncertainty.
The non-parametric DE is effectively optimized in a similar fashion.
In contrast, NLL-optimized PU networks have a home-turf advantage on these datasets since the parametric variance is explicitly optimized to account for the present heteroscedastic aleatoric uncertainty.
\ours is the only non-parametric approach that accounts for the presence of this kind of uncertainty.
While the results look similar, the underlying mechanisms are fundamentally different.
On the one hand explicit prediction of the uncertainty, on the other hand implicit modeling via distribution matching. 
Accompanying quantitative evaluations can be found in appendix \ref{appendix:toyEval}.
To collect further evidence that \ours approximates the true ground truth uncertainty $\sigma_{\rm true}$ appropriately, we fit it to `noisy line' toy datasets in appendix \ref{appendix:toyEval}. Both large and small $\sigma_{\rm true}$ values are correctly matched, indicating that \ours is not just adding an uncertainty offset but flexibly spreads/contracts its sub-networks as intended.
In the following, we substantiate the corroborative results of \ours on toy data by an empirical study on 1D standard datasets and an application to a modern object detection network.


\subsection{Standard 1D regression datasets}
\label{sec:uci}

Next, we study standard regression datasets, extending the dataset selection in \cite{gal2016dropout} by adding four additional datasets: `diabetes', `abalone', `california', and `superconduct'. Table \ref{tab:uci_data_preproc} in appendix \ref{appendix:uci_eval} provides details on dataset sources, preprocessing and basic statistics.
Apart from train- and test-data results, we study regression performance and uncertainty quality \textit{under data shift}. Such distributional changes and uncertainty quantification are closely linked since the latter ones are rudimentary ``self-assessment'' mechanisms that help to judge model reliability. These judgements gain importance for model inputs that are \textit{structurally different} from the training data.

\begin{figure}[thb]
    \centering 
    \includegraphics[trim=50 180 50 60, clip, width=1.0\textwidth]{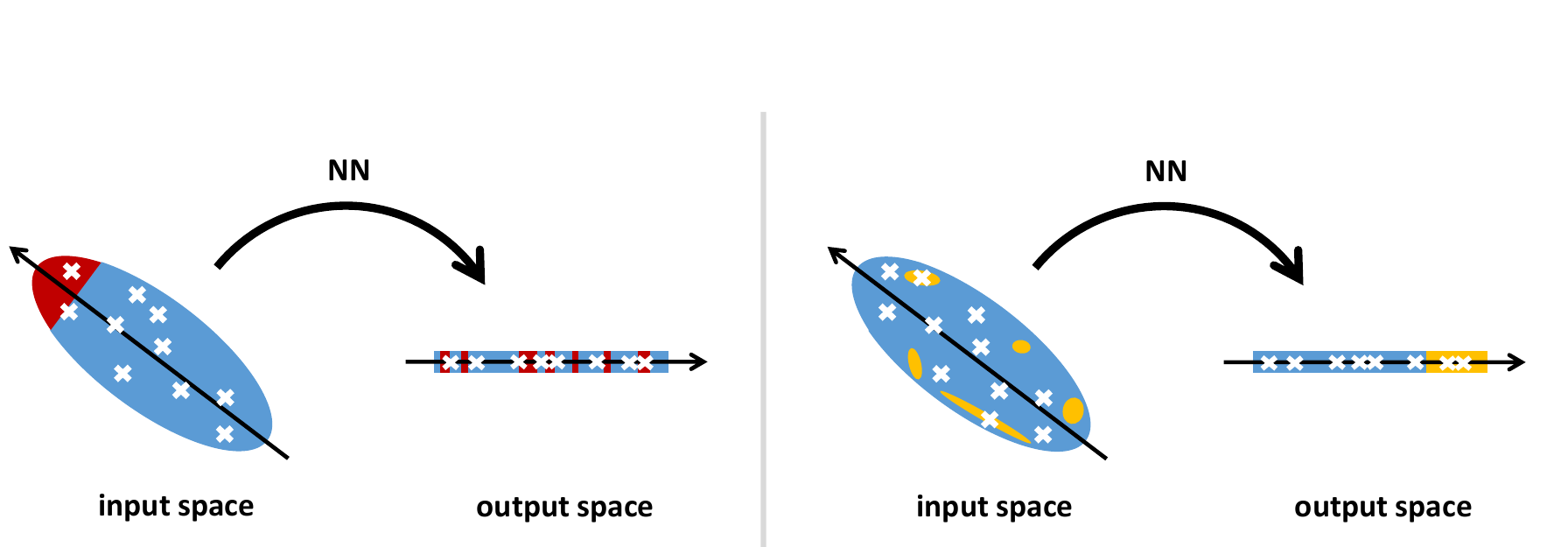}
    \caption{Scheme of two non-\iid splits: a PCA-based split in input space (left) and label-based split in output space (right). While datasets appear to be convex here, they are (most likely) not in reality.}
    \label{fig:pca_label_split}
\end{figure}

\paragraph{Data splits} 
Natural candidates for such non-\iid splits are splits along the main directions of data in input and output space, respectively. Here, we consider 1D regression tasks. Therefore, output-based splits are simply done on a scalar label variable (see Fig.\ \ref{fig:pca_label_split}, right). We call such a split \textit{label-based} (for a comparable split, see, \egnows, \cite{foong2019between}). In input space, the first component of a principal component analysis (PCA) provides a natural direction (see Fig.\ \ref{fig:pca_label_split}, left). 
Projecting the data points onto this first PCA-axis yields the scalar values the \textit{PCA-split} is based on.
Note that these projections are only considered for data splitting, they are not used for model training. Splitting data along such a direction in input or output space in, \egnows, $10$ equally large chunks, creates $2$ \textit{outer} data chunks and $8$ \textit{inner} data chunks. Training a model on $9$ of these chunks such that the remaining chunk for evaluation is an inner chunk is called data \textit{interpolation}. If the remaining test chunk is an outer chunk, it is data \textit{extrapolation}. For example, for labels running from $0$ to $1$, (label-based) extrapolation testing would consider only data with a label larger $0.9$, while training would be performed on the smaller label values. We introduce this distinction as extrapolation is expected to be considerably more difficult than `bridging' between feature combinations that were seen during training.

\begin{figure}[bth]
    \centering
    \includegraphics[trim=10 10 40 40, clip, width=1.0\textwidth]{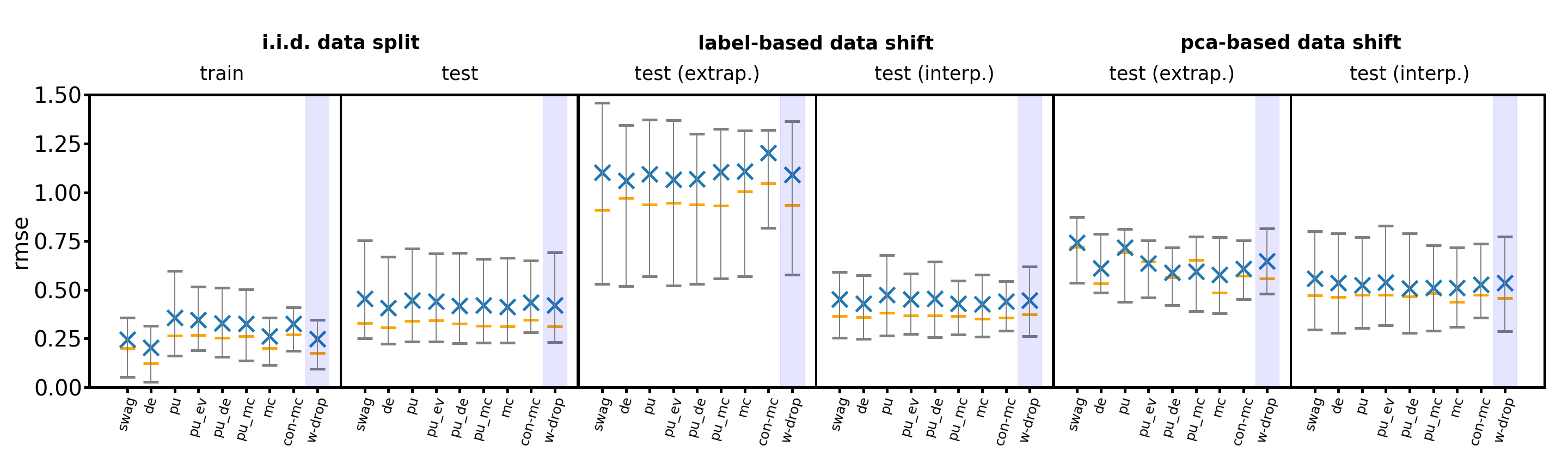} 
    \includegraphics[trim=40 40 40 40, clip, width=1.0\textwidth]{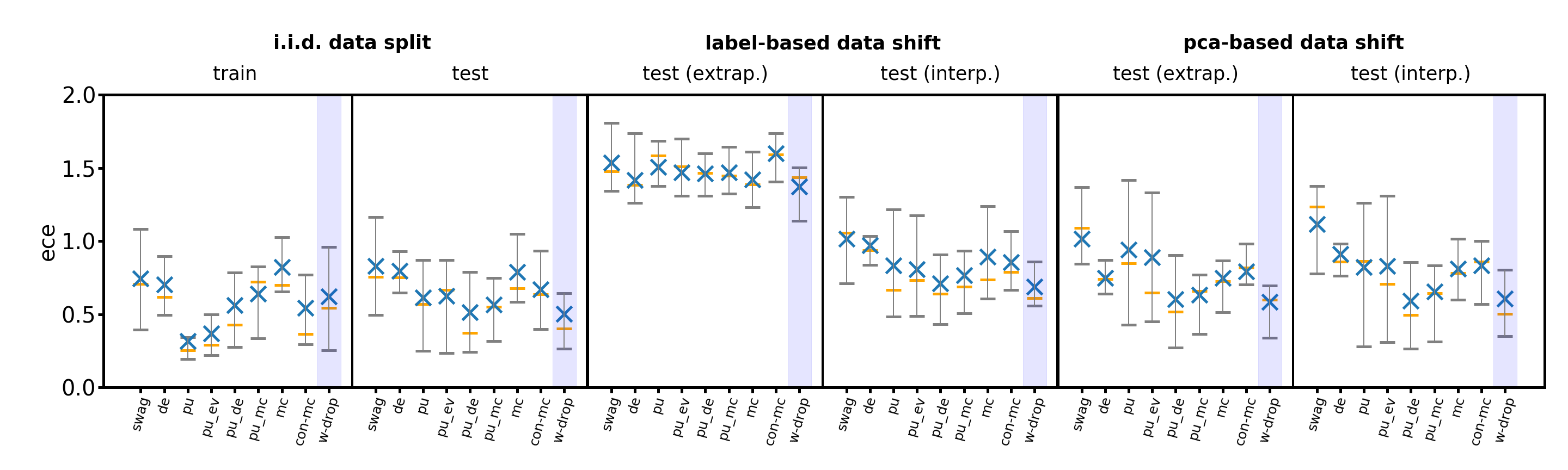} 
    \caption{Root-mean-square errors (RMSEs ($\downarrow$), top row) and expected calibration errors (ECEs ($\downarrow$), bottom row) of different uncertainty methods under \iid conditions (first and second panel in each row) and under various kinds of data shift (third to sixth panel in each row, see text for details). \ours (light blue background) is compared to 8 benchmark approaches. Each blue cross is the mean over 14 1D regression datasets. Orange line markers indicate median values. The gray vertical bars reach from the 25\% quantile (bottom horizontal line) to the 75\% quantile (top horizontal line).}
    \label{fig:uci_ece}
    \label{fig:uci_rmse}
\end{figure}

\begin{figure*}[t]
    \centering
    \includegraphics[trim=40 10 40 40, clip, width=1.0\textwidth]{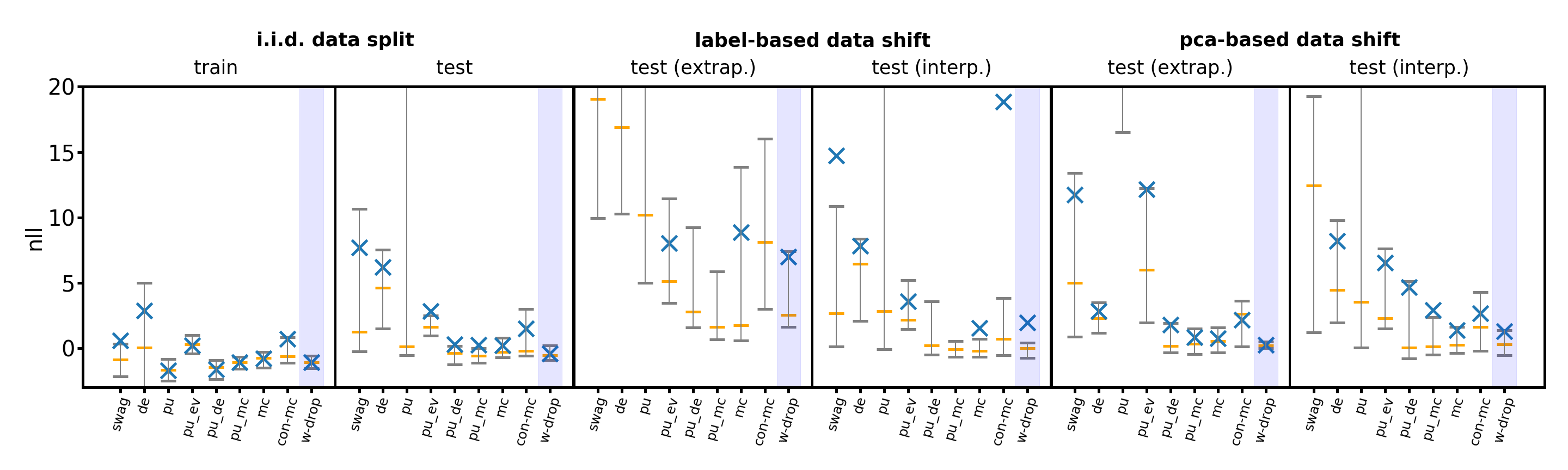} 
    \includegraphics[trim=40 40 40 40, clip, width=1.0\textwidth]{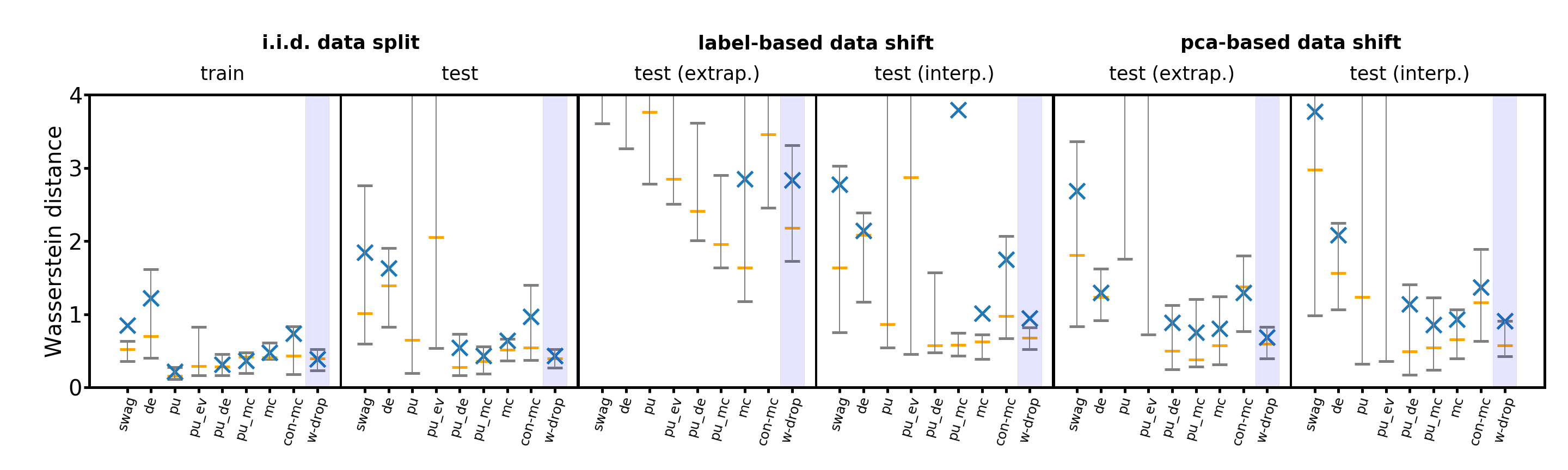} 
    \caption{Negative log-likelihoods (NLLs ($\downarrow$), top row) and Wasserstein distances ($\downarrow$\,, bottom row) of different uncertainty methods under \iid conditions (first and second panel in each row) and under various kinds of data shift (third to sixth panel in each row, see text for details). \ours (light blue background) is compared to 8 benchmark approaches. Each blue cross is the mean over ECE values from 14 standard regression datasets. Orange line markers indicate median values. The gray vertical bars reach from the 25\% quantile (bottom horizontal line) to the 75\% quantile (top horizontal line).}
    \label{fig:uci_nll_ws}
\end{figure*}

More general information on training and dataset-dependent modifications to the experimental setup are relegated to the technical appendix \ref{appendix:experimentalSetup}. 
The presented results are obtained as follows:
for each of the 14 standard datasets, we calculate (for each uncertainty method) the per-dataset scores: RMSE, mean NLL, ECE and WS. 
To improve statistical significance, these scores are $5$- or $10$-fold cross-validated, \ie averages across a respective number of folds. 
Given the (fold-averaged) per-dataset scores for all 14 standard datasets, we calculate and visualize their mean and median values as well as quantile intervals 
(see Fig.\ \ref{fig:uci_rmse} and Fig.\ \ref{fig:uci_nll_ws}). 
For high-level summaries of the results on in-data and out-of-data test sets please refer to Tab.\ \ref{tab:uci_id_summary} and Tab.\ \ref{tab:uci_ood_summary}, respectively.
While the mean values characterize the average behavior of the uncertainty methods, the displayed 75\% quantiles indicate how well methods perform on the more challenging datasets. 
A small 75\% quantile value thus hints at \textit{consistent} stability of an uncertainty mechanism across a variety of tasks.


\begin{table}[bth]
    \centering
    \caption{Regression performance (RMSE) and uncertainty quality (NLL, ECE, WS) of \ours and various uncertainty benchmarks. \ours yields the best uncertainty scores while providing a competitive RMSE value. Each number is the average across 14 standard 1D (test) datasets. The figures in this table correspond to the blue crosses in the second columns of Fig.\ \ref{fig:uci_rmse} and Fig.\ \ref{fig:uci_nll_ws}, respectively. See text for further details.}
    \begin{tabular}{lrrrr}
    \toprule
  {unc. method} & {RMSE ($\downarrow$)} & {NLL ($\downarrow$)} & {ECE ($\downarrow$)} & {WS ($\downarrow$)} \\
\bottomrule
  \addlinespace[0.5em]
         SWAG & $0.456$ & $7.695$ & $0.828$ & $1.847$ \\
DE & $\mathbf{0.407}$ & $6.184$ & $0.796$ & $1.628$ \\
PU & $0.447$ & $1.47\cdot 10^{7}$ & $0.614$ & $2.10\cdot 10^{6}$ \\
PU-EV & $0.442$ & $2.838$ & $0.626$ & $49.180$ \\
PU-DE & $0.418$ & $0.307$ & $0.515$ & $0.542$ \\
PU-MC & $0.420$ & $0.250$ & $0.565$ & $0.433$ \\
MC & $0.412$ & $0.190$ & $0.788$ & $0.643$ \\
CON-MC & $0.436$ & $1.513$ & $0.669$ & $0.964$ \\
W-Dropout & $0.421$ & $\mathbf{-0.428}$ & $\mathbf{0.501}$ & $\mathbf{0.430}$ \\
\bottomrule
    \end{tabular}
    \label{tab:uci_id_summary}
\end{table}

\begin{table}[bth]
    \centering
    \caption{Out-of-data analysis of \ours and various uncertainty benchmarks. Regression performance (RMSE) and uncertainty quality (NLL, ECE, WS) are displayed. As for in-domain test data, \ours outperforms the other uncertainty methods without sacrificing regression quality. Each number is obtained by two-fold averaging: firstly, across two types of out-of-data test sets (label-based and PCA-based splits) and secondly, across 14 standard 1D datasets. The figures in this table are based on the blue crosses in the last four columns of Fig.\ \ref{fig:uci_rmse} and Fig.\ \ref{fig:uci_nll_ws}, respectively. See text for further details.}
    \begin{tabular}{lrrrr}
    \toprule
  {unc. method} & {RMSE ($\downarrow$)} & {NLL ($\downarrow$)} & {ECE ($\downarrow$)} & {WS ($\downarrow$)} \\
\bottomrule
  \addlinespace[0.5em]
SWAG & $0.641$ & $27.602$ & $1.138$ & $3.818$ \\
DE & $0.599$ & $14.055$ & $0.988$ & $2.554$ \\
PU & $0.632$ & $1.50\cdot 10^{7}$ & $0.968$ & $1.55\cdot 10^{6}$ \\
PU-EV & $0.611$ & $6.290$ & $0.941$ & $44.447$ \\
PU-DE & $0.594$ & $1448.391$ & $0.783$ & $5.892$ \\
PU-MC & $0.591$ & $397.022$ & $0.823$ & $3.215$ \\
MC & $\mathbf{0.589}$ & $2.330$ & $0.923$ & $1.207$ \\
CON-MC & $0.621$ & $13.820$ & $0.963$ & $2.109$ \\
W-Dropout & $0.615$ & $\mathbf{2.287}$ & $\mathbf{0.763}$ & $\mathbf{1.203}$ \\
\bottomrule
    \end{tabular}
    \label{tab:uci_ood_summary}
\end{table}

\paragraph{Regression quality} First, we consider regression performance, see Tab.\ \ref{tab:uci_id_summary} and the first two panels in the top row of Fig.\ \ref{fig:uci_ece}.
Averaging the RMSE values across the 14 datasets yields almost identical test results for all uncertainty methods (see Tab.\ \ref{tab:uci_id_summary}). 
On training data (Fig.\ \ref{fig:uci_ece}, first panel in top row) in contrast, we find the parametric methods to exhibit larger train data RMSEs which could be due to NLL optimization favoring to adapt variance rather than mean. However, this regularizing NLL training comes along with a smaller generalization gap, leading to competitive test RMSEs (see Tab.\ \ref{tab:uci_id_summary} and the second panel in the top row of Fig.\ \ref{fig:uci_ece}). 
\ours is on a par with the benchmark approaches, \ie our optimization objective does not lead to degraded regression quality.
Next, we investigate model performance under data shift, 
visualized in the third to sixth panel in the top row of Fig.\ \ref{fig:uci_ece}. For interpolation setups (fourth and sixth panel), regression quality is comparable between all methods. As expected, performances under these data shifts are (slightly) worse compared to those on \iid test sets. The more challenging extrapolation setups (third and fifth panel) amplify the deterioration in performance across all methods. 
Again, \ours yields competitive RMSE values (see also Tab.\ \ref{tab:uci_ood_summary}).


\paragraph{Expected calibration errors} Fig.\ \ref{fig:uci_ece} (bottom row) provides average ECE values of the outlined uncertainty methods under i.i.d.\ conditions (first and second panel), under label-based data shifts (third and fourth panel) and under PCA-based data shifts (fifth and sixth panel). 
On training data, PU performs best, followed by PU-EV and all other methods.
Interestingly, both SWAG and \ours show a relatively broad range of ECE values on the various training datasets.
This could be interpreted as a form of over-estimation of the present uncertainty and for \ours this effect occurs on mostly smaller datasets with lower data variability.
However, looking at the \iid test results (Tab.\ \ref{tab:uci_id_summary} and second panel in the bottom row of Fig.\ \ref{fig:uci_rmse}) we find 
\ours to provide the lowest averaged ECE (Tab.\ \ref{tab:uci_id_summary}), followed by the PU-based (implicit) ensembles of PU-DE and PU-MC.
The calibration quality of \ours is moreover the most consistent one across the datasets as can be seen from its small 75\% quantile value (Fig.\ \ref{fig:uci_ece}, second panel in bottom row).

Looking at the stability \wrt data shift, \ienows, extra- and interpolation based on label-split or PCA-split, again \ours reaches the smallest calibration errors (followed by PU-DE and PU-MC, see Tab.\ \ref{tab:uci_ood_summary}). Regarding the 75\% quantiles, \ours consistently provides one of the best results 
on all out-of-data (OOD) test sets.

\paragraph{Negative log-likelihoods}
For the unbounded NLL (see Tab.\ \ref{tab:uci_id_summary} and the top row of Fig.\ \ref{fig:uci_nll_ws}), the results are more widely distributed compared to the (bounded) ECE values. 
\ours reaches the smallest mean value on \iid test sets, followed by MC and PU-MC (Tab.\ \ref{tab:uci_id_summary}). 
The mean NLL value of PU is above the upper plot limit in Fig.\ \ref{fig:uci_nll_ws} (second panel in the upper row) indicating a rather weak stability of this method. 
On PCA-interpolate and PCA-extrapolate test sets (Fig.\ \ref{fig:uci_nll_ws}, last two panels in the upper row), MC, PU-MC and \ours networks perform best. On label-interpolate and label-extrapolate test sets, only MC and \ours networks are in first place when considering average values, followed by PU-EV. 
The mean NLLs of many other approaches are above the upper plot limit.
Averaging all these OOD results in Tab.\ \ref{tab:uci_ood_summary}, we find \ours to provide the overall smallest NLL values, narrowly followed by MC.
Note that median results are not as widely spread and PU-DE, MC, PU-MC and \ours perform comparably well. 
These qualitative differences between mean and median behavior indicate that most methods perform poorly `once in a while'. 
A noteworthy observation as \textit{stability across a variety of data shifts and datasets} can be seen as a crucial requirement for an uncertainty method. 
\ours models yield high stability in that sense \wrt NLL.


\paragraph{Wasserstein distances}
Studying Wasserstein distances, we again observe the smallest scores on test data for \oursnows, followed by PU-MC and PU-DE (see Tab.\ \ref{tab:uci_id_summary} and the second panel in the bottom row of Fig.\ \ref{fig:uci_nll_ws}). 
While PU provides the best WS value on training data, its generalization behavior is less stable: on test data, its mean and 75\% quantile take high values beyond the plot range. Under data shift (Tab.\ \ref{tab:uci_ood_summary} and third to sixth panel in bottom row of Fig.\ \ref{fig:uci_nll_ws}), 
\ours and MC are in the lead, CON-MC and DE follow on ranks three and four.
On label-based data shifts, MC and \ours outperform all other methods by a significant margin when considering average values. As for NLL, we find the mean values for PU-DE and PU-MC to be significantly above their respective median values indicating again weaknesses \wrt the stability of parametric methods. Here as well, not only good \textit{average} results, but also consistency over the datasets and splits, is a hallmark of Wasserstein dropout.

\begin{figure}[ht]
    \centering
    \includegraphics[width=0.65\textwidth]{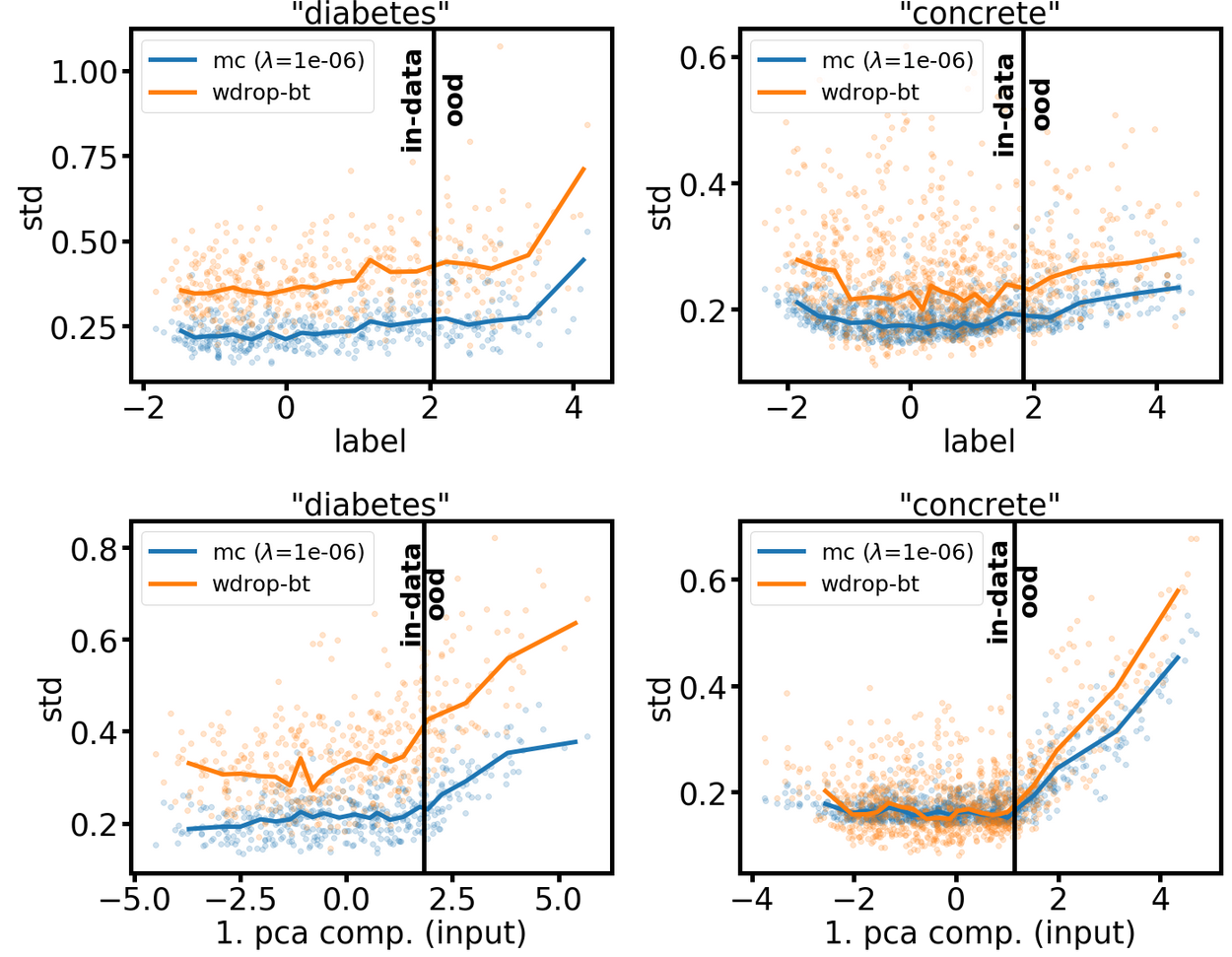}
    \caption{Extrapolation behavior of \ours (orange) and MC dropout (blue). Two extrapolation ``directions'' (rows) and two datasets (columns) are considered. The vertical bar in each panel separates training data (left) from out-of-data (OOD, right). Scatter points show the predicted standard deviation for individual data points. The colored solid lines show averages over points in equally-sized bins and reflect the expected growth of epistemic uncertainty in the OOD-region. For details on the data splits and extrapolations please refer to subsection \ref{sec:uci} and appendix \ref{appendix:uci_eval}.}
    \label{fig:wdrop_ood_behavior}
\end{figure}

\paragraph{Epistemic uncertainty} Summarizing these evaluations on 1D regression datasets, we find \ours to yield better and more stable uncertainty estimates than the state-of-the-art methods of PU-DE and PU-MC.
We moreover observe advantages for \ours under PCA- and label-based data shifts.
These results suggest that \ours induces uncertainties which increase under data shift, \ienows, it approximately models \textit{epistemic uncertainty}. This conjecture is supported by Fig.\ \ref{fig:wdrop_ood_behavior} that visualizes the uncertainties of MC dropout (blue) and \mbox{\oursnows} (orange) for transitions from in-data to out-of-data. As expected, these shifts lead to increased (epistemic) uncertainty for MC dropout.
This holds true for \ours that behaves highly similar under data shift indicating that it ``inherits" this ability \hide{the ability to model epistemic uncertainty} from MC dropout: both approaches match sub-networks to training data and these sub-networks ``spread'' when leaving the training data distribution. 
Since \ours models heteroscedastic, \ie input-dependent, aleatoric uncertainty, we notice a higher variability of its uncertainties in Fig.\ \ref{fig:wdrop_ood_behavior} compared to the ones of
MC dropout.

For further (visual) inspections of uncertainty quality, see the residual-uncertainty scatter plots in appendix \ref{appendix:res_error}. A reflection on NLL and comparisons of the different uncertainty measures on 1D regression datasets can be found in appendix \ref{appendix:uci_eval}.

\begin{table}[ht] 
\centering
\caption{Study of worst-case scenarios for different uncertainty methods: \ours (W-Drop), PU-DE and PU-MC are compared to the ideal Gaussian case for \iid and non-\iid data splits. Uncertainty quality in these scenarios is quantified by the expected tail loss at the $99\%$ quantile (ETL$_{0.99}$). Each mean and max value is taken over the ETLs of $110$ models trained on $15$ different datasets.}\label{tab:etl_stats}
\vspace{0.8em}
\setlength\tabcolsep{1.8pt} 
\begin{tabular}{r l r r r r}
\toprule
  {measure} & {split} & {$\mathcal{N}(0,1)$} & {W-Drop} & {PU-DE} & {PU-MC} \\
\bottomrule
  \addlinespace[0.5em]
mean ETL$_{0.99}$  & \iid   & $2.89$ & $4.68$          & $7.77$    & $6.24$    \\
max ETL$_{0.99}$   & \iid   & $3.01$ & $8.86$          & $31.14$   & $91.75$   \\
\addlinespace[0.5em]                                   
mean ETL$_{0.99}$  & label  & $2.89$ & $7.28$          & $86.79$   & $44.00$   \\
max ETL$_{0.99}$   & label  & $3.01$ & $93.55$         & $2267.93$ & $1224.27$ \\
\addlinespace[0.5em]                                   
mean ETL$_{0.99}$  & pca    & $2.89$ & $5.93$          & $9.78$    & $8.62$    \\
max ETL$_{0.99}$   & pca    & $3.01$ & $18.35$         & $64.13$   & $93.49$   \\  
  \addlinespace[0.4em]
\bottomrule
  \addlinespace[-0.4em]
\end{tabular}
\end{table}


\paragraph{Expected tail loss} 
For both toy and standard regression datasets, we calculate the expected tail loss at the 99\% quantile (ETL$_{0.99}$) on test data. Doing this for all trained networks yields a total of $110$ ETL$_{0.99}$ values per uncertainty method when including cross-validation.
As a tail measure, the ETL$_{0.99}$ evaluates a specific aspect of the distribution of uncertainty estimates. Studying such a property is useful if the uncertainty estimate distribution as a whole is appropriate, as measured \eg by the ECE. We thus restrict the ETL$_{0.99}$ analysis to the three methods that provide the best ECE values, namely PU-MC, PU-DE and \oursnows.
The mean and maximum values of their ETL$_{0.99}$'s are reported in Table \ref{tab:etl_stats}.
While none of these methods gets close to the ideal ETL$_{0.99}$'s of the desired $\mathcal{N}(0,1)$ Gaussian, \ours networks exhibit significantly less pronounced tails and therefore higher stability compared to PU-MC and PU-DE.
This holds true over all considered test sets. Deviations from standard normal increase from the \iid train-test split over the PCA-based train-test split to the label-based one.
We attribute the lower stability of PU-DE to the nature of the PU networks that compose the ensemble, although their inherent instability (see Table \ref{tab:apx_eval_uci} in appendix \ref{appendix:uci_eval}) is largely suppressed by ensembling.
Considering the tail of the distribution of the prediction residuals $\vert r_i\vert $, however, reveals that regular\-iza\-tion of PU by ensembling might not work in every single case. It is then unlikely that larger ensemble are able to fully cure this instability issue. Regularizing PU by applying dropout (PU-MC) leads to only mild improvement. \ours networks in contrast encode uncertainty into the structure of the entire network thus yielding improved stability compared to parametric approaches. 
Further analysis shows that the large normalized residuals $r_i=(\mu_i-y_i)/\sigma_i$, which cause the large $\text{ETL}_{0.99}$ values, correspond (on average) to large absolute errors $(\mu_i-y_i)$.\footnote{They are (on average) not due to small absolute residuals $\ll 1$ that go along with even smaller uncertainty estimates.} This underpins the practical relevance of the ETL analysis, as large absolute errors are more harmful than small ones in many contexts, \eg when detecting traffic participants.


\paragraph{Dependencies between uncertainty measures} All uncertainty-related measures (NLL, ECE, WS, ETL) relate predicted uncertainties to actually occurring model residuals. Each of them putting emphasize on different aspects of the considered samples: NLL is biased towards well-performing models, ECE measures deviations within quantile ranges, Wasserstein distance resolves distances between normalized residuals and ETL focuses on distribution tails. The empirically observed dependencies between WS and ECE are visualized in Fig.\ \ref{fig:ece_ws_uci_data}.
Additionally to WS and ECE, we consider Kolmogorov-Smirnov (KS) distances \citep{stephens1974edf} on normalized residuals in Fig.\ \ref{fig:corrs_uncertainty_measures} in appendix \ref{appendix:unc_measures}.

While all these scores are expectably correlated, noteworthy deviations from ideal correlation occur.
Therefore, we advocate for uncertainty evaluations based on various measures to avoid overfitting to a specific formalization of uncertainty. The top panel of Fig.\ \ref{fig:corrs_uncertainty_measures} reflects the higher sensitivity of the Wasserstein distance compared to ECE: we observe two ``slopes", the first one corresponds to models that overestimate uncertainties, \ienows, $\sigma_{\bthetaTilde} > \vert \mu_{\bthetaTilde} - y_i\vert $ on average. In these scenarios, WS is typically below $1$ as $1$ would be the WS distance between a delta distribution at zero (corresponding to $\sigma_{\bthetaTilde} \to \infty$) and the expected $\mathcal{N}(0,1)$ Gaussian. 
The second ``slope" contains models that underestimate uncertainties, \ienows, $\sigma_{\bthetaTilde} < \vert \mu_{\bthetaTilde} - y_i\vert $. WS is not bounded in these scenarios and is thus---unlike ECE---able to resolve differences between any two uncertainty estimators.


\subsection{Application to object regression}\label{sec:squeezeDet}
\begin{table}[ht] 
\centering
\caption{Basic statistics of the harmonized object detection datasets. Dataset size and number of annotated objects are reported for train data (first two columns) and test data (last two columns). For details on dataset harmonization, see text and references therein.} \label{tab:od_dataset_stats}
\vspace{0.8em}
\setlength\tabcolsep{6pt} 
\begin{tabular}{l r r r r}
\toprule
 & \multicolumn{2}{c}{train} & \multicolumn{2}{c}{test} \\
\cmidrule(lr){2-3} \cmidrule(lr){4-5}
dataset & {\# images} & {\# objects} & {\# images} & {\# objects}\\
\bottomrule
\addlinespace[0.5em]
KITTI	  & 3.622  & 15.254  & 3.387  &  12.673 \\
SynScapes &	19.998 & 906.827 & 4.998  &	 226.390  \\
A2D2	  & 22.731 & 121.320 & 4.186  &  36.544   \\
Nightowls &	30.064 & 50.225  & 6.595  &	 10.766   \\
NuImages  & 58.803 & 410.462 & 14.377 &  97.014   \\
BDD100k	  & 69.281 & 843.963 & 9.919  &	 123.752  \\
\addlinespace[0.5em]
\bottomrule
\end{tabular}
\end{table}

After studying toy and standard regression datasets, we turn towards the challenging task of object detection (OD), namely the SqueezeDet model \citep{wu2017squeezedet}, a fully convolutional neural network. 
First, we adopt the \ours objective to SqueezeDet (see the following paragraph). 
Next, we introduce the six considered OD datasets and sketch central technical aspects of training and inference.
Since OD networks are often employed in open-world applications (like autonomous vehicles or drones), they likely encounter various types of concept shifts during operations. 
In such novel scenarios, well-calibrated ``self-assessment'' capabilities help to foster safe functioning. 
We therefore evaluate Wasserstein-SqueezeDet not only in-domain but on corrupted and augmented test data as well as on other object detection datasets
(see last paragraphs of this subsection). 


\paragraph{Architecture}
SqueezeDet takes an RGB input image and predicts three quantities: (i) $2D$ bounding boxes for detected objects (formalized as a $4D$ regression task), (ii) a confidence score for each predicted bounding box and (iii) the class of each detection.
Its architecture is as follows: First, a sequence of convolutional layers extracts features from the input image. Next, dropout with a drop rate of $p=0.5$ is applied to the final feature representations.
Another convolutional layer, the ConvDet layer, finally estimates prediction candidates. In more detail, SqueezeDet predictions are based on so-called anchors, initial bounding boxes with prototypical shapes.
The ConvDet layer computes for each such anchor a confidence score, class scores and offsets to the initial position and shape.
The final prediction outputs are obtained by applying a non-maximum-suppression (NMS) procedure to the prediction candidates.
The original loss of SqueezeDet is the sum of three terms. It  reads \mbox{$L_{\rm SqueezeDet} = L_{\rm regres} + L_{\rm conf} + L_{\rm class}$} with the bounding box regression loss  $L_{\rm regres}$, a confidence-score loss $L_{\rm conf}$ and the object-classification loss $L_{\rm class}$. Our modification of the learning objective is restricted to the L2 regression loss: 
\begin{flalign}
    L_{\rm regres} = \frac{\lambda_{\rm bbox}}{N_{\rm obj}}&\sum_{i=1}^{W} \sum_{j=1}^{H} \sum_{k=1}^{K} \sum_{\xi \in \{x,y,w,h\}} I_{ijk} \bigg[({\delta \xi}_{ijk} - \delta \xi_{ijk}^G)^2 \bigg] 
\end{flalign}
with ${\delta \xi}_{ijk}$ and $\delta \xi_{ijk}^G$ being estimates and ground truth expressed in coordinates relative to the $k$-th anchor at grid point $(i,j)$ where $\xi\in\{x,y,w,h\}$. 
See \cite{wu2017squeezedet} for descriptions of all other loss parameters. Applying \ours component-wise to this 4D regression problem yields

\begin{align*}
    L_{\rm regres, W} = \frac{\lambda_{\rm bbox}}{N_{\rm obj}} \sum_{i=1}^{W} \sum_{j=1}^{H} \sum_{k=1}^{K} \sum_{\xi \in \{x,y,w,h\}} I_{ijk} & \bigg[\mathcal{W}(\xi_{ijk}) \bigg]\,,
\end{align*}
where
\begin{align*}
    \mathcal{W}(\xi_{ijk}) = &
    \left(\mu_{\delta \xi_{ijk}} - \delta\xi_{ijk}^G\right)^2 \\
    + & \left(\sqrt{\sigma_{\delta\xi_{ijk}}^2} -\sqrt{\left(\mu_{\delta\xi_{ijk}} - \delta\xi_{ijk}^G\right)^2 + \sigma_{\delta\xi_{ijk}}^2}\right)^2
\end{align*}
with $\mu_{\delta \xi_{ijk}} = \frac{1}{L} \sum_{l=1}^L \delta \xi_{ijk}^{(l)}$ being the sample mean and $\sigma_{\delta \xi_{ijk}}^2 = \frac{1}{L} \sum_{l=1}^L (\delta \xi_{ijk}^{(l)} - \mu_{\delta \xi_{ijk}})^2$ being the sample variance over $L$ dropout predictions $\delta \xi_{ijk}^{(l)}$ for $\xi\in\{x,y,w,h\}$.


\paragraph{Datasets} We train SqueezeDet networks on six traffic scene datasets: KITTI \citep{Geiger2012CVPR}, SynScapes \citep{wrenninge2018synscapes}, A2D2 \citep{geyer2020a2d2}, Nightowls \citep{neumann2018nightowls}, NuImages (NuScenes) \citep{caesar2020nuscenes} and BDD100k \citep{yu2018bdd100k}. 
They differ from each other in dataset size (the large BDD100k dataset contains almost 20 times more images than the small KITTI dataset, see Table \ref{tab:od_dataset_stats}), time of day (Nightowls comprises only nighttime images) and data acquisition (SynScapes is simulation-based). For further information on the datasets, see Table \ref{tab:od_datasets_general_info} in appendix \ref{appendix:squeeze}. 
We employ image sizes of $672 \times 384$ and rescale all datasets (except for KITTI\footnote{For KITTI, we crop images in x-direction to avoid strong distortions due to its high aspect ratio. In y-direction, only a minor upscaling is applied.}) accordingly. 
To facilitate cross-dataset model evaluations (see paragraphs on OOD analyses in this section), we group the various object classes of the six datasets into three main categories: `pedestrian', `cyclist' and `vehicle' (see Table \ref{tab:od_class_mapping} in appendix \ref{appendix:squeeze} for the object class mapping). Some static or rare object classes are discarded.\\


\paragraph{Technical aspects} We compare 
MC-SqueezeDet, \ienows, standard SqueezeDet with activated dropout at inference, with W-SqueezeDet that uses \mbox{\ours} instead of the original MSE regression loss.
All models are trained for $300{,}000$ mini-batches of size $20$.
After training, we keep dropout active and compute $50$ forward passes for each test image. The detections from all forward passes are clustered using k-means \citep{bishop2006pattern}.\footnote{Using the density-based clustering technique HDBSCAN \citep{campello2013density} yields comparable results especially \wrt the relative ordering of the methods.} The number of clusters is chosen for each image to match the average number of detections across the 50 forward passes. Each cluster is summarized by its mean detection and standard deviation. To ensure meaningful statistics, we discard clusters with 4 or less detections. The cluster means are matched with ground truth. We exclude predictions from the evaluation if their IoU with ground truth is $\leq 0.1$. For each dataset, SqueezeDet's maximum number of detections is chosen proportionally to the average number of ground truth objects per image.
\begin{table}[ht] 
\centering
\caption{Regression performance and uncertainty quality of SqueezeDet-type networks on KITTI data. W-SqueezeDet (W-SqzDet) is compared with the default MC-SqueezeDet (MC-SqzDet).
The values of NLL, ECE and WS are aggregated across their respective four dimensions, for details see appendix \ref{appendix:squeeze} and Table \ref{tab:squeezedet_details} therein.} \label{tab:od_test}
\vspace{0.8em}
\setlength\tabcolsep{6pt} 
\begin{tabular}{l r r r r}
\toprule
{~ } & \multicolumn{2}{c}{train} & \multicolumn{2}{c}{test} \\
\cmidrule(lr){2-3} \cmidrule(lr){4-5}
{measure} & {MC-SqzDet} & {W-SqzDet} & {MC-SqzDet} & {W-SqzDet}\\
\bottomrule
\addlinespace[0.5em]
mIoU $(\uparrow)$ & $\textbf{0.705}$ & $0.691$ & $\textbf{0.695}$ & $0.694$ \\
RMSE $(\downarrow)$ & $\textbf{8.769}$ & $9.832$ & $14.666$ & $\textbf{14.505}$ \\
\addlinespace[0.5em]
NLL $(\downarrow)$ & $8.497$ & $\textbf{2.770}$ & $25.704$ & $\textbf{6.309}$ \\
ECE $(\downarrow)$ & $0.615$ & $\textbf{0.193}$ & $0.825$ & $\textbf{0.433}$ \\
WS $(\downarrow)$ & $1.421$ & $\textbf{0.315}$ & $2.831$ & $\textbf{0.900}$ \\
ETL$_{0.99}$  & $22.358$ & $8.853$ & $42.101$ & $18.223$ \\
\addlinespace[0.5em]
\bottomrule
\end{tabular}
\end{table}

\begin{figure}[t]
    \centering
    \includegraphics[width=0.7\textwidth]{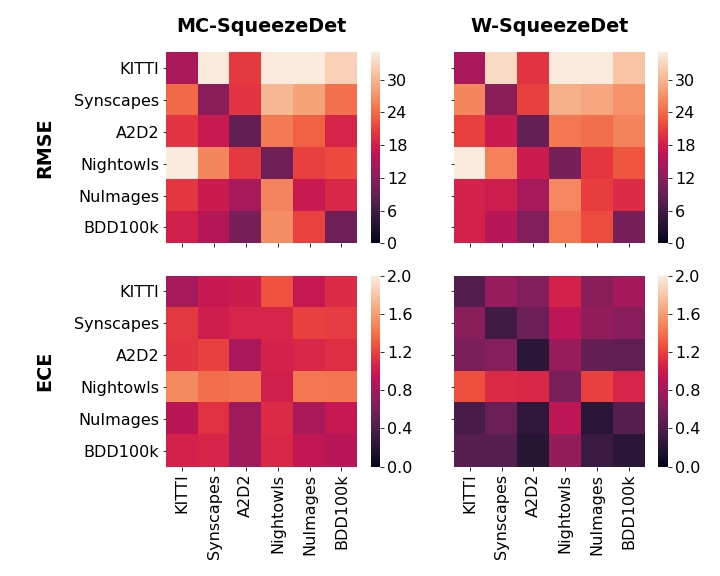} 
     \caption{In-data and out-of-data evaluation of MC-SqueezeDet (lhs) and W-SqueezeDet (rhs) on six OD datasets. We consider regression quality (RMSE, top row) and uncertainty quality (ECE, bottom row). For each heatmap entry, the row label refers to the training dataset, the column label to the test dataset. Thus, diagonal matrix elements are in-data evaluations, non-diagonal elements are OOD analyses. W-SqueezeDet provides substantially smaller ECE values both in-data and out-of-data.} 
    \label{fig:od_in-data_ood_mc_w}
\end{figure}

\paragraph{In-data evaluation} To assess model performance, we report the mean intersection over union (mIoU) and RMSE (in pixel space) between predicted bounding boxes and matched ground truths. The quality of the uncertainty estimates is measured by (coordinate-wise) NLL, ECE, WS and ETL. Table \ref{tab:od_test} shows a summary of our results on train and test data for the KITTI dataset. The results for NLL, ECE, WS and ETL have been averaged across the 4 regression coordinates. 
MC-SqueezeDet (abbreviated as MC-SqzDet) and W-SqueezeDet \mbox{(W-SqzDet)} show comparable regression results in terms of RMSE and mIoU, with slight advantages for MC-SqueezeDet. 
Considering uncertainty quality, we find substantial advantages for W-SqueezeDet across all evaluation measures. 
These advantages are due to the estimation of heteroscedastic aleatoric uncertainty during training (see also the test statistics `trajectories' during training for BDD100k in Fig.\ \ref{fig:od_stats_during_training_bdd} in appendix \ref{appendix:squeeze}). 
The test RMSE and ECE values of all six OD datasets are visualized as diagonal elements in Fig.\ \ref{fig:od_in-data_ood_mc_w}. The (mostly) `violet' RMSE diagonals for MC-SqueezeDet and W-SqueezeDet (top row of Fig.\ \ref{fig:od_in-data_ood_mc_w}) again indicate comparable regression performances. Datasets are ordered by size from small (top) to large (bottom). The large NuImages test set occurs to be the most challenging one. Regarding ECE (bottom row of Fig.\ \ref{fig:od_in-data_ood_mc_w}), W-SqueezeDet performs consistently stronger, see the `violet' W-SqueezeDet diagonal (smaller values) and the `red' MC-SqueezeDet diagonal (higher values). These findings qualitatively resemble those on the standard regression datasets and indicate that \ours works well on a modern application-scale network.

To analyze how well these OD uncertainty mechanisms function on test data that is structurally different from training data, we consider two types of 
out-of-data analyses in the following:
first, we study \mbox{SqueezeDet} models that are trained on one OD dataset and evaluated on the test sets of the remaining five OD datasets. A rather `semantic' OOD study as features like object statistics and scene composition vary between training and OOD test sets. Second, we consider networks that are trained on one OD dataset and evaluated on corrupted versions (defocus blur, Gaussian noise) of the respective test set, thus facing changed `low-level' features, \ie less sharp edges due to blur and textures overlayed with pixel noise, respectively.

\paragraph{Out-of-data evaluation on other OD datasets} We train one SqueezeDet on each of the six OD datasets and evaluate each of these models on the test sets of the remaining 5 datasets. The resulting OOD regression scores and OOD ECE values are visualized as off-diagonal elements in Fig.\ \ref{fig:od_in-data_ood_mc_w} for MC-SqueezeDet (left column) and W-SqueezeDet (right column). Since datasets are ordered by size (a rough proxy to dataset complexity), the upper triangular matrix corresponds to cases in which the evaluation dataset is especially challenging (``easy to hard"), while the lower triangular matrix subsumes easier test sets compared to the respective \iid test set (``hard to easy"). Accordingly, we observe (on average) lower RMSE values in the lower triangular matrix for both SqueezeDet variants. The ECE values of W-SqueezeDet are once more smaller (`violet') compared to MC-SqueezeDet (`red'). The ECE diagonal of W-SqueezeDet is visually more pronounced compared to the one of MC-SqueezeDet since uncertainty calibration is effectively optimized during the training of W-SqueezeDet. The Nightowls dataset causes a cross-shaped pattern, indicating that neither transfers of Nightowls models to other datasets nor transfers from other models to Nightowls work well. This behavior can be understood as the feature distributions of Nightowls' nighttime images diverge from the (mostly) daytime images of the other datasets. The high uncertainty quality of W-SqueezeDet is underpinned by the evaluations of NLL and WS (see Fig.\ \ref{fig:od_id-ood_nll_ws} and text in appendix \ref{appendix:squeeze}).


\begin{table}[ht] 
\centering
\caption{Out-of-data evaluation of MC-SqueezeDet (MC-SqzDet) and W-SqueezeDet \mbox{(W-SqzDet)} on distorted OD datasets. Each model is trained on the original dataset and evaluated on two modified versions of the respective test set: a blurred one (first two columns) and a noisy one (last two columns), see text for details. We report the expected calibration error (ECE) and find W-SqueezeDet to perform better than MC-SqueezeDet on most datasets.}\label{tab:od_ood_corrupted}
\vspace{0.8em}
\setlength\tabcolsep{6pt} 
\begin{tabular}{l r r r r}
\toprule
 & \multicolumn{2}{c}{defocus blur} & \multicolumn{2}{c}{Gaussian noise} \\
\cmidrule(lr){2-3} \cmidrule(lr){4-5}
dataset & {MC-SqzDet} & {W-SqzDet} & {MC-SqzDet} & {W-SqzDet}\\
\bottomrule
\addlinespace[0.5em]
KITTI & $\mathbf{1.034}$ & $1.082$ & $\mathbf{1.021}$ & $1.084$ \\
SynScapes & $1.081$ & $\mathbf{0.503}$ & $0.941$ & $\mathbf{0.910}$ \\
A2D2 & $0.921$ & $\mathbf{0.295}$ & $1.143$ & $\mathbf{0.617}$ \\
Nightowls & $1.067$ & $\mathbf{0.803}$ & $0.992$ & $\mathbf{0.682}$ \\
NuImages & $0.908$ & $\mathbf{0.332}$ & $\mathbf{0.760}$ & $0.849$ \\
BDD100k & $1.012$ & $\mathbf{0.390}$ & $0.833$ & $\mathbf{0.633}$ \\
\addlinespace[0.5em]
\bottomrule
\label{tab:od_blurred_noisy_testsets}
\end{tabular}
\end{table}

\paragraph{Out-of-data evaluation on corrupted datasets}
In contrast to the analysis above, we now focus on `non-semantic' data shifts due to technical distortions. For each test set, we generate a blurred and a noisy version.\footnote{We employ the \texttt{imgaug} library (\url{https://github.com/aleju/imgaug}) and apply defocus blur (severity of ``1") and additive Gaussian noise (\iid per pixel, drawn from the distribution $\mathcal{N}(0,20)$), respectively.} Two examples of these transformations can be found in Fig.\ \ref{fig:od_images_blur_noise} in appendix \ref{appendix:squeeze}. In accordance with previous results, W-SqueezeDet provides smaller ECE values compared to MC-SqueezeDet on most blurred and noisy test sets (see Table \ref{tab:od_blurred_noisy_testsets}). 
We observe a less substantial deterioration of uncertainty quality for blurring compared to adding pixel noise, possibly because the latter one more strongly affects short-range pixel correlations that the networks rely on.


\section{Conclusion}
\label{sec:disc}
The prevailing approaches to uncertainty quantification rely on parametric uncertainty estimates by means of a dedicated network output. 
In this work, we propose a novel type of uncertainty mechanism, \textit{Wasserstein dropout}, that quantifies (aleatoric) uncertainty in a purely non-parametric manner:
by revisiting and newly assembling core concepts from existing dropout-based uncertainty methods, we construct distributions of randomly drawn sub-networks that closely approximate the actual data distributions.
This is achieved by a natural extension of the Euclidean metric ($L_2$-loss) for points to the 2-Wasserstein metric for distributions. In the limit of vanishing distribution width, \mbox{\ie vanishing} uncertainty, both metrics coincide. 
Assuming Gaussianity and making a bootstrap approximation, the metric can be replaced by a compact loss objective affording stable training.
To the best of our knowledge, \ours is the first \textit{non-parametric} method to model aleatoric uncertainty in neural networks. It outperforms the ubiquitous parametric approaches, as, \egnows, shown by our comparison to deep ensembles (PU-DE).

An extensive additional study of uncertainties under data shift further reveals advantages of \ours models compared to deep ensembles (PU-DE) and parametric models combined with dropout (PU-MC): the Wasserstein-based technique still provides (on average) better calibrated uncertainty estimates while coming along with a higher stability across a variety of datasets and data shifts. 
In contrast, we find parametric uncertainty estimation (PU) to be prone to instabilities that are only partially cured by the regularizing effects of explicit or implicit (dropout-based) ensembling (PU-DE, PU-MC).
With respect to worst-case scenarios, W-dropout networks are by a large margin better than either PU-DE or PU-MC.
This makes \ours especially suitable 
for safety-critical applications like automated driving or medical diagnosis where (even rarely occurring) inadequate uncertainty estimates might lead to injuries and damage.
Furthermore, while our 
theoretical derivation focuses on
aleatoric uncertainty, 
the presented distribution-shift experiments 
suggest that \ours is also 
able to capture 
epistemic uncertainty. Finding a theoretical explanation for that is subject of future research.

With respect to computational demands, \ours is roughly equivalent to MC dropout (MC) and, in fact, could be used as a drop-in replacement for the latter.
While $L$-fold sampling of sub-networks increases the training complexity, we observe an increase of training time that is significantly below $L$ in our implementation.
Inference is performed in the same way for both methods and thus also their run-time complexities are equivalent.
In comparison to deep ensembles, \oursnows's use of a single network reduces requirements on training and storage at the expense of multiple forward passes during inference.
This property is shared with MC and approaches exist to reduce the prediction cost, for instance last-layer MC allows sampling-free inference (see also \cite{postels2019sampling}).

In addition to the toy and 1D regression experiments, SqueezeDet is selected as a representative of large-scale object detection networks.
We find the above mentioned properties of Wasserstein dropout to carry over to Wasserstein-SqueezeDet, namely the enhanced uncertainty quality and its increased stability under different types of data shifts. 
At the same time observed performance losses are minimal.
Overall, our experiments on SqueezeDet show that \ours scales to larger networks relevant for practical applications.

Taking a step back, the idea to ``migrate" from single-point modeling to full distributions is a very general one and can be applied to a variety of tasks. Replacing, \mbox{\egnows, Gaussians} with Dirichlet distributions makes an application to classification conceivable, where \cite{malinin2018predictive} employ parametric (Dirichlet) distributions to quantify uncertainty.
Conceptually, our findings suggest that distribution modeling based on \textit{sampling} generalizes better
compared to parameterized counterparts. An observation that might find applications far outside the scope of uncertainty quantification.


\section*{Acknowledgments}
The research of J.\ Sicking and M.\ Akila was funded by the German Federal Ministry for Economic Affairs and Energy within the project ``KI Absicherung – Safe AI for Automated Driving''. Said authors would like to thank the consortium for the successful cooperation. The work of T.\ Wirtz was funded by the German Federal Ministry of Education and Research, ML2R - no. 01S18038B. S.\ Wrobel contributed as part of the Fraunhofer Center for Machine Learning within the Fraunhofer Cluster for Cognitive Internet Technologies. The work of A.~Fischer was supported by the Deutsche Forschungsgemeinschaft (DFG, German Research Foundation) under Germany’s Excellence Strategy – EXC-2092 \textsc{CaSa} – 390781972.


\bibliography{references}

\clearpage

\appendix
{\LARGE
{Supplementary Material} 
}
\vspace{0.3cm}

This part accompanies our paper ``\textit{Wasserstein Dropout}'' and provides further in-depth information.
Large parts of the empirical evaluation on toy data and standard regression datasets can be found in section \ref{appendix:empiricalStudy}, including details on the datasets, more granular evaluations and additional toy data experiments.
Details on the object detection datasets and supplementary evaluations of SqueezeDet are located in subsection \ref{appendix:squeeze}.
As \ours exhibits the hyper-parameters $p$ (drop rate) and $L$ (sample size), we test various values in section \ref{appendix:hyper}, finding no strong correlation between result and parameter choices.
We close with a discussion on the relation between uncertainty measures and their respective sensitivity in section \ref{appendix:unc_measures}.


\section{Extension to the empirical study}
\label{appendix:empiricalStudy}

Complementing the evaluation sketched in the body of the paper, section \ref{sec:experiments}, we provide more details on the training setup and benchmark approaches in the following subsection.
Further information on the toy dataset experiments can be found in subsection \ref{appendix:toyEval}.
The same holds for the 1D regression experiments in subsection \ref{appendix:uci_eval}, which we extend by evaluations on dataset level that were skipped in the main text.
A close look at the predicted uncertainties (per method) on these datasets is given via scatter plots in subsection \ref{appendix:res_error}.
Details on OD dataset preprocessing and SqueezeDet results are found in the last subsection.


\subsection{Experimental setup}
\label{appendix:experimentalSetup}

The experimental setup used for the toy data and 1D regression experiments is presented in two parts: first, technical details of the benchmark approaches we compare with and second, a description of the neural networks and training procedures we employ. 

For MC dropout, we choose the regularization coefficient $\lambda$ by grid search on the set $\lambda \in \{0, 10^{-6}, 10^{-5}, 10^{-4}, 10^{-3}, 10^{-2}\}$ and find $\lambda = 10^{-6}$ to provide the best overall results for the 1D regression datasets. A variant of MC dropout that optimizes its layer-specific drop rates during training is Concrete dropout (CON-MC): all its initial drop rates are set to $p_{\rm initial} = 0.1$. The hyper-parameters $wr = l^2/(\tau N)$ and $dr = 2/N$ are determined by the number of training datapoints $N$, prior length scale $l = 10^{-3}$ and $\tau(N) \in [10^{-3}, 2]$ that decreases monotonically with N. For PU and PU-EV networks, we ensure positivity constraints using softplus \citep{relu3} and optimize Gaussian NLL and t-distribution NLL, respectively. 
The regularization coefficient of PU-EV is set to $\lambda = 10^{-2}$, determined by a grid search considering the parameter range $\lambda \in \{10^{-4}, 10^{-3}, 10^{-2}, 0.1, 0.5\}$. 
For SWAG, we start to estimate the low-rank Gaussian proxy (rank $r = 20$) for the NN weight distribution after training for $n/2$ epochs, with $n$ being the total number of training epochs.

We categorize the toy and 1D regression datasets as follows: small datasets \{toy-hf, yacht, diabetes, boston, energy, concrete, wine-red\}, large datasets \{toy-noise, abalone, kin8nm, power, naval, california, superconduct, protein\} and very large datasets \{year\}. For small datasets, NNs are trained for $1{,}000$ epochs using mini-batches of size $100$. All results are 10-fold cross validated. For large datasets, we train for $150$ epochs and apply 5-fold cross validation. We keep this large-dataset setting for the very large `year' dataset but increase mini-batch size to $500$.

All experiments are conducted on \texttt{Core Intel(R) Xeon(R) Gold 6126} CPUs and\\ \texttt{NVidia Tesla V100} GPUs.
Conducting the described experiments with cross validation on one CPU takes $20\,h$ for toy data, $130\,h$ for 1D regression datasets and approximately $100\,h$ for object regression on the GPU.


\subsection{Toy datasets: systematic evaluation and further experiments}\label{appendix:toyEval}
The toy-noise and toy-hf datasets are sampled from \mbox{$f_{\rm noise}(x) \sim \mathcal{N}(0,\exp(-0.02\ x^2))$} for \mbox{$x \in [-15,15]$} and $f_{\rm hf}(x) = 0.25\,x^2 - 0.01\,x^3 + 40\,\exp(-(x + 1)^2/\,200)\,\sin(3\,x)$ for \mbox{$x \in [-15,20]$}, respectively. Standard normalization is applied to input and output values.
Detailed evaluations of the considered uncertainty methods on these datasets are given in Table \ref{tab:apx_eval_toy}.

\begin{figure}[th]
    \centering
    \includegraphics[width=1.00\textwidth]{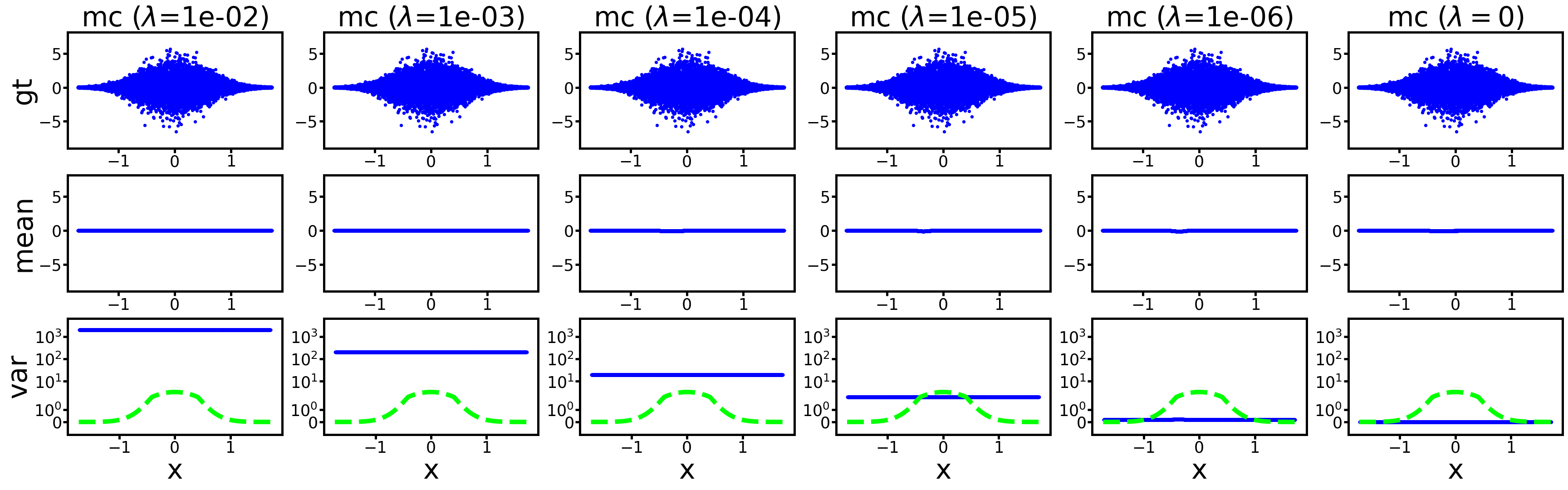} 
     \caption{MC dropout and aleatoric uncertainty. The regularization parameter $\lambda$ of MC dropout allows to model the average (homoscedastic) noise level of a dataset. As the regularizer is not input-dependent, it does not capture the x-dependency of the noise level, \ie the heteroscedasticity of the dataset, see third row.} 
    \label{fig:toy_modulated_mc_lambda}
\end{figure}

\begin{table*}[!th]
\centering
\caption{Regression performance and uncertainty quality of networks with different uncertainty mechanisms. All scores are calculated on the test set of toy-hf and toy-noise, respectively.}\label{tab:apx_eval_toy}
\vspace*{0.8em}
\setlength\tabcolsep{6.0pt} 
\begin{tabular}{l l r r r r r}
\toprule
\textbf{measure} &  \textbf{dataset} & \textbf{swag} & \textbf{de} & \textbf{pu} & \textbf{pu-ev} & \textbf{pu-de} \\
\bottomrule
\addlinespace[0.5em]
RMSE ($\downarrow$) & toy-hf    & $0.696$    & $0.660$            & $0.691$  & $0.691$     & $0.690$  \\
NLL ($\downarrow$)  & toy-hf    & $85.331$   & $52.444$           & $-0.098$ & $1.855$     & $-0.100$ \\
ECE ($\downarrow$)  & toy-hf    & $1.472$    & $1.584$            & $0.548$  & $0.500$     & $0.524$  \\
WS ($\downarrow$)   & toy-hf    & $9.043$    & $7.413$            & $0.233$  & $0.242$     & $0.243$  \\
\midrule
\addlinespace[0.4em]
RMSE ($\downarrow$) & toy-noise & $1.006$    & $1.006$            & $1.006$  & $1.006$     & $1.006$  \\
NLL ($\downarrow$)  & toy-noise & $6934.498$ & $1.14\cdot 10^{4}$ & $-0.374$ & $1.555$     & $-0.374$ \\
ECE ($\downarrow$)  & toy-noise & $1.541$    & $1.642$            & $0.062$  & $0.098$     & $0.084$  \\
WS ($\downarrow$)   & toy-noise & $63.760$   & $83.590$           & $0.028$  & $0.064$     & $0.048$  \\
\addlinespace[0.5em]
\bottomrule
\end{tabular}
\begin{tabular}{l l r r r r}
\addlinespace[0.5em]
\toprule
\textbf{measure} &  \textbf{dataset} & \textbf{pu-mc} & \textbf{con-mc} & \textbf{mc} & \textbf{w-drop}\\
\bottomrule
\addlinespace[0.5em]
RMSE ($\downarrow$)  & toy-hf    & $0.694$  & $0.701$            & $0.696$  & $0.678$  \\
NLL ($\downarrow$)   & toy-hf    & $-0.083$ & $17.616$           & $13.370$ & $-0.055$ \\
ECE ($\downarrow$)   & toy-hf    & $0.544$  & $1.380$            & $1.352$  & $0.428$  \\
WS ($\downarrow$)    & toy-hf    & $0.233$  & $4.356$            & $3.830$  & $0.222$  \\
\midrule
\addlinespace[0.4em]
RMSE ($\downarrow$)  & toy-noise & $1.007$  & $0.995$            & $1.006$  & $1.013$  \\
NLL ($\downarrow$)   & toy-noise & $-0.370$ & $6.57\cdot 10^{4}$ & $1.723$  & $-0.330$ \\
ECE ($\downarrow$)   & toy-noise & $0.066$  & $1.730$            & $0.645$  & $0.107$  \\
WS ($\downarrow$)    & toy-noise & $0.030$  & $5.03\cdot 10^{4}$ & $0.693$  & $0.054$  \\
\addlinespace[0.5em]
\bottomrule
\end{tabular}

\end{table*}
To illustrate the capabilities and limitations of MC dropout regarding the modeling of aleatoric uncertainty, we consider the toy-noise dataset again and systematically vary MC's regularization parameter $\lambda$ (see Fig.\ \ref{fig:toy_modulated_mc_lambda}, $\lambda$ decreases from left to right). As MC dropout's uncertainty estimates contain an additive constant term proportional to $\lambda$, tuning this parameter allows to model the \textit{average} aleatoric uncertainty (the ideal $\lambda$ in Fig.\ \ref{fig:toy_modulated_mc_lambda} is between $\lambda = 10^{-6}$ and $\lambda = 10^{-5}$). Input dependencies of noise (heteroscedasticity) can however not be incorporated, \ie even an optimized $\lambda$ causes systematic over- and under-estimations of the data uncertainty in many cases. This is in contrast to \oursnows.

\begin{figure}[th]
    \centering
    \includegraphics[width=0.75\textwidth]{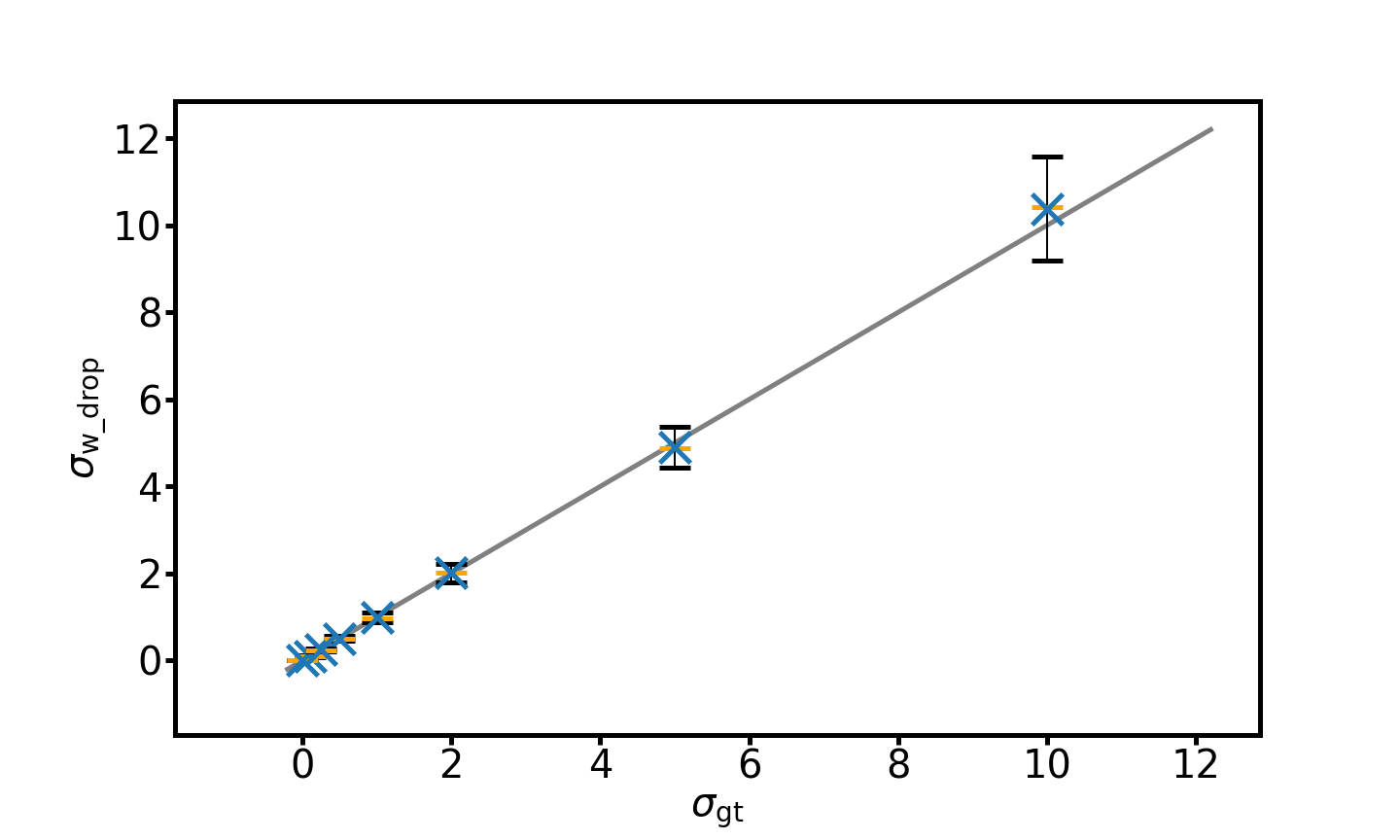} 
     \caption{Standard deviation $\sigma_{\rm w\text{-}drop}$ of W-dropout (y-axis) when fitted to a toy dataset with ground truth standard deviation $\sigma_{\rm gt}$ (x-axis, see text for details). The bisecting line is shown in gray. While $\sigma_{\rm w\text{-}drop}$ exhibits fluctuations (black `whiskers' at $10\%$ and $90\%$ quantile), it provides on average accurate estimates of the ground truth uncertainty. Both mean value (blue cross) and median value (orange bar) of $\sigma_{\rm w\text{-}drop}$ are close to the bisector.} 
    \label{fig:noisy_line_w-drop}
\end{figure}

Having shown that \ours can approximate input-dependent data uncertainty appropriately (see Fig.\ \ref{fig:subnets_toydata}), we now analyze its ability to match ground truth uncertainties $\sigma_{\rm true}$ more systematically. Therefore, we fit a ‘noisy line’ toy dataset that is given by $(x_i,y_i)$ with $x_i \sim \mathcal{U}(-1,1)$ and $y_i \sim \mathcal{N}(0, \sigma_{\mathrm{true}})$. The ground truth standard deviations take the values $\sigma_{\rm true} = 0, 0.1, 0.2, 0.5, 1, 2, 5, 10$. Fig.\ \ref{fig:noisy_line_w-drop} emphasizes that \ours provides accurate uncertainty estimates for both small and large noise levels. Minor x-dependent fluctuations (see `whiskers' in Fig.\ \ref{fig:noisy_line_w-drop}) decrease monotonically with $\sigma_{\rm true}$.



\subsection{Standard regression datasets: systematic evaluation}\label{appendix:uci_eval}

An overview on the 1D regression datasets providing basic statistics and information on preprocessing is given in Table \ref{tab:uci_dataset_stats}. Evaluations of RMSE, NLL, ECE and WS on dataset level can be found in Table \ref{tab:uci_eval_per_dataset}.

\begin{table}[th] 
\centering
\caption{Details on 1D regression datasets. Ground truth (gt) is partially preprocessed to match the 1D regression setup.}\label{tab:uci_data_preproc}
\vspace{0.8em}
\setlength\tabcolsep{6pt} 
\begin{tabular}{l r r l p{4.5cm}}
\toprule
  {dataset} & {\# features} & {\# data points} & {source} & {remarks} \\
\bottomrule
  \addlinespace[0.5em]
yacht       &  6    &  308      & UCI & \\
diabetes    &  7    &  442      & StatLib & \\
boston      &  13   &  506      & StatLib & \\
energy      &  8    &  768      & UCI & {Only the ``cooling load" label is used.} \\
concrete    &  8    &  1030     & UCI & \\
wine-red    &  11   &  1599     & UCI & \\
abalone     &  7    &  4176     & UCI & {The categorical input feature ``sex" is omitted.} \\
kin8nm      &  8    &  8192     & Delve & \\
power       &  4    &  9568     & UCI & \\
naval       &  16   &  11934    & UCI & {Only the ``turbine" label is used.} \\
california  &  8    &  20640    & StatLib & \\
superconduct&  81   &  21263    & UCI & \\
protein     &  9    &  45730    & UCI & \\
year        &  90   &  515345   & UCI & \\
\bottomrule
  \addlinespace[0.8em]
\end{tabular}
\label{tab:uci_dataset_stats}
\end{table}

\begin{table*}
\centering
\caption{Regression performance and uncertainty quality of networks with different uncertainty mechanisms. The scores are calculated on the test sets of 14 standard regression datasets.}\label{tab:apx_eval_uci}
\vspace{0.8em}
\begin{adjustbox}{max width=0.77\textwidth} 
\begin{tabular}{l l r r r r r r r r r}
\toprule
\textbf{measure} &  \textbf{dataset} & \textbf{swag} & \textbf{de} & \textbf{pu} & \textbf{pu-ev} & \textbf{pu-de} & \textbf{pu-mc} & \textbf{con-mc} & \textbf{mc} & \textbf{w-drop}\\
\bottomrule
\addlinespace[0.5em]
RMSE $(\downarrow)$ & yacht & $0.055$ & $0.042$ & $0.062$ & $0.051$ & $0.049$ & $0.076$ & $0.103$ & $0.074$ & $0.075$ \\
NLL $(\downarrow)$ & yacht & $-3.010$ & $-3.560$ & $-2.370$ & $-0.438$ & $-3.017$ & $-2.881$ & $-2.059$ & $-2.202$ & $-2.489$ \\
ECE $(\downarrow)$ & yacht & $0.757$ & $0.653$ & $0.863$ & $0.890$ & $1.087$ & $1.093$ & $1.010$ & $1.093$ & $1.113$ \\
WS $(\downarrow)$ & yacht & $0.470$ & $0.379$ & $0.668$ & $1.228$ & $0.544$ & $0.560$ & $0.481$ & $0.536$ & $0.537$ \\
\addlinespace[0.4em]
\hline
\addlinespace[0.4em]
RMSE $(\downarrow)$ & diabetes & $0.990$ & $0.920$ & $0.816$ & $0.832$ & $0.790$ & $0.773$ & $0.792$ & $0.843$ & $0.886$ \\
NLL $(\downarrow)$ & diabetes & $61.217$ & $7.629$ & $1712.520$ & $11.505$ & $7.423$ & $2.302$ & $3.427$ & $6.014$ & $2.356$ \\
ECE $(\downarrow)$ & diabetes & $1.456$ & $0.942$ & $1.353$ & $1.370$ & $0.945$ & $0.854$ & $0.975$ & $1.108$ & $0.909$ \\
WS $(\downarrow)$ & diabetes & $7.704$ & $1.717$ & $19.552$ & $19.182$ & $1.936$ & $1.230$ & $1.630$ & $2.229$ & $1.265$ \\
\addlinespace[0.4em]
\hline
\addlinespace[0.4em]
RMSE $(\downarrow)$ & boston & $0.352$ & $0.322$ & $0.339$ & $0.349$ & $0.329$ & $0.287$ & $0.317$ & $0.310$ & $0.318$ \\
NLL $(\downarrow)$ & boston & $10.739$ & $6.116$ & $553.188$ & $4.003$ & $3.590$ & $0.050$ & $-0.371$ & $-0.040$ & $-0.287$ \\
ECE $(\downarrow)$ & boston & $1.170$ & $0.882$ & $1.294$ & $1.240$ & $0.876$ & $0.644$ & $0.636$ & $0.688$ & $0.624$ \\
WS $(\downarrow)$ & boston & $2.987$ & $1.593$ & $9.233$ & $5.288$ & $1.490$ & $0.658$ & $0.493$ & $0.674$ & $0.581$ \\
\addlinespace[0.4em]
\hline
\addlinespace[0.4em]
RMSE $(\downarrow)$ & energy & $0.097$ & $0.077$ & $0.118$ & $0.166$ & $0.106$ & $0.079$ & $0.107$ & $0.080$ & $0.075$ \\
NLL $(\downarrow)$ & energy & $-1.048$ & $0.628$ & $5.882$ & $1.751$ & $-2.007$ & $-2.056$ & $-1.703$ & $-1.815$ & $-2.047$ \\
ECE $(\downarrow)$ & energy & $0.567$ & $0.602$ & $1.037$ & $0.819$ & $0.502$ & $0.705$ & $0.637$ & $0.819$ & $0.648$ \\
WS $(\downarrow)$ & energy & $0.672$ & $0.723$ & $2.183$ & $1.508$ & $0.374$ & $0.379$ & $0.332$ & $0.424$ & $0.339$ \\
\addlinespace[0.4em]
\hline
\addlinespace[0.4em]
RMSE $(\downarrow)$ & concrete & $0.258$ & $0.241$ & $0.263$ & $0.260$ & $0.244$ & $0.237$ & $0.270$ & $0.235$ & $0.248$ \\
NLL $(\downarrow)$ & concrete & $2.726$ & $4.899$ & $34.402$ & $2.496$ & $0.208$ & $-0.978$ & $-0.672$ & $-0.890$ & $-0.868$ \\
ECE $(\downarrow)$ & concrete & $0.672$ & $0.649$ & $0.871$ & $0.763$ & $0.528$ & $0.384$ & $0.395$ & $0.431$ & $0.400$ \\
WS $(\downarrow)$ & concrete & $1.279$ & $1.402$ & $3.217$ & $2.605$ & $0.773$ & $0.230$ & $0.256$ & $0.267$ & $0.253$ \\
\addlinespace[0.4em]
\hline
\addlinespace[0.4em]
RMSE $(\downarrow)$ & wine-red & $0.931$ & $0.835$ & $0.824$ & $0.817$ & $0.771$ & $0.783$ & $0.748$ & $0.784$ & $0.807$ \\
NLL $(\downarrow)$ & wine-red & $10.343$ & $1.352$ & $1.41\cdot 10^{5}$ & $10.201$ & $3.578$ & $11.142$ & $2.485$ & $1.962$ & $0.830$ \\
ECE $(\downarrow)$ & wine-red & $0.946$ & $0.488$ & $0.765$ & $0.816$ & $0.456$ & $0.541$ & $0.677$ & $0.664$ & $0.549$ \\
WS $(\downarrow)$ & wine-red & $2.487$ & $0.717$ & $35.052$ & $55.450$ & $0.813$ & $0.914$ & $1.154$ & $1.014$ & $0.536$ \\
\addlinespace[0.4em]
\hline
\addlinespace[0.4em]
RMSE $(\downarrow)$ & abalone & $0.798$ & $0.701$ & $0.653$ & $0.657$ & $0.654$ & $0.636$ & $0.636$ & $0.684$ & $0.718$ \\
NLL $(\downarrow)$ & abalone & $13.289$ & $31.002$ & $2.05\cdot 10^{8}$ & $2.486$ & $-0.073$ & $-0.111$ & $9.206$ & $1.017$ & $0.256$ \\
ECE $(\downarrow)$ & abalone & $1.181$ & $1.276$ & $0.252$ & $0.250$ & $0.238$ & $0.270$ & $1.082$ & $0.496$ & $0.402$ \\
WS $(\downarrow)$ & abalone & $3.149$ & $4.624$ & $3.07\cdot 10^{7}$ & $286.448$ & $0.157$ & $0.141$ & $2.630$ & $0.726$ & $0.462$ \\
\addlinespace[0.4em]
\hline
\addlinespace[0.4em]
RMSE $(\downarrow)$ & kin8nm & $0.259$ & $0.246$ & $0.272$ & $0.276$ & $0.253$ & $0.269$ & $0.313$ & $0.266$ & $0.261$ \\
NLL $(\downarrow)$ & kin8nm & $-0.393$ & $1.905$ & $-0.142$ & $1.274$ & $-0.866$ & $-0.631$ & $-0.612$ & $-0.678$ & $-0.855$ \\
ECE $(\downarrow)$ & kin8nm & $0.453$ & $0.677$ & $0.462$ & $0.362$ & $0.205$ & $0.561$ & $0.223$ & $0.491$ & $0.151$ \\
WS $(\downarrow)$ & kin8nm & $0.552$ & $1.202$ & $0.629$ & $0.405$ & $0.185$ & $0.334$ & $0.203$ & $0.306$ & $0.080$ \\
\addlinespace[0.4em]
\hline
\addlinespace[0.4em]
RMSE $(\downarrow)$ & power & $0.228$ & $0.219$ & $0.225$ & $0.227$ & $0.221$ & $0.228$ & $0.234$ & $0.226$ & $0.226$ \\
NLL $(\downarrow)$ & power & $1.682$ & $12.696$ & $-0.921$ & $0.870$ & $-1.024$ & $-1.007$ & $-0.470$ & $-0.720$ & $-0.788$ \\
ECE $(\downarrow)$ & power & $0.755$ & $1.075$ & $0.154$ & $0.176$ & $0.135$ & $0.172$ & $0.415$ & $0.656$ & $0.264$ \\
WS $(\downarrow)$ & power & $1.306$ & $2.943$ & $0.127$ & $0.193$ & $0.071$ & $0.098$ & $0.520$ & $0.393$ & $0.318$ \\
\addlinespace[0.4em]
\hline
\addlinespace[0.4em]
RMSE $(\downarrow)$ & naval & $0.250$ & $0.030$ & $0.163$ & $0.194$ & $0.165$ & $0.169$ & $0.345$ & $0.118$ & $0.100$ \\
NLL $(\downarrow)$ & naval & $-0.198$ & $-2.815$ & $-2.358$ & $0.327$ & $-1.405$ & $-1.464$ & $-0.522$ & $-0.701$ & $-1.479$ \\
ECE $(\downarrow)$ & naval & $1.230$ & $0.898$ & $0.677$ & $0.887$ & $1.235$ & $0.760$ & $0.378$ & $1.233$ & $0.888$ \\
WS $(\downarrow)$ & naval & $0.615$ & $0.487$ & $0.357$ & $0.931$ & $0.606$ & $0.454$ & $0.297$ & $0.632$ & $0.483$ \\
\addlinespace[0.4em]
\hline
\addlinespace[0.4em]
RMSE $(\downarrow)$ & california & $0.444$ & $0.430$ & $0.674$ & $0.514$ & $0.475$ & $0.549$ & $0.456$ & $0.436$ & $0.448$ \\
NLL $(\downarrow)$ & california & $0.813$ & $7.305$ & $-0.494$ & $1.266$ & $-0.612$ & $-0.560$ & $1.441$ & $-0.236$ & $-0.213$ \\
ECE $(\downarrow)$ & california & $0.469$ & $0.829$ & $0.248$ & $0.212$ & $0.251$ & $0.314$ & $0.625$ & $0.650$ & $0.251$ \\
WS $(\downarrow)$ & california & $0.745$ & $1.965$ & $0.181$ & $306.628$ & $0.162$ & $0.194$ & $0.989$ & $0.357$ & $0.312$ \\
\addlinespace[0.4em]
\hline
\addlinespace[0.4em]
RMSE $(\downarrow)$ & superconduct & $0.305$ & $0.290$ & $0.340$ & $0.341$ & $0.326$ & $0.346$ & $0.345$ & $0.318$ & $0.310$ \\
NLL $(\downarrow)$ & superconduct & $-0.242$ & $3.045$ & $-0.558$ & $0.622$ & $-1.340$ & $-1.169$ & $-0.065$ & $-0.341$ & $-0.954$ \\
ECE $(\downarrow)$ & superconduct & $0.346$ & $0.515$ & $0.151$ & $0.181$ & $0.204$ & $0.243$ & $0.381$ & $0.916$ & $0.183$ \\
WS $(\downarrow)$ & superconduct & $0.451$ & $1.126$ & $0.204$ & $0.250$ & $0.129$ & $0.141$ & $0.569$ & $0.487$ & $0.161$ \\
\addlinespace[0.4em]
\hline
\addlinespace[0.4em]
RMSE $(\downarrow)$ & protein & $0.615$ & $0.575$ & $0.723$ & $0.696$ & $0.701$ & $0.666$ & $0.654$ & $0.610$ & $0.610$ \\
NLL $(\downarrow)$ & protein & $0.560$ & $4.314$ & $0.117$ & $1.515$ & $-0.132$ & $-0.080$ & $3.149$ & $0.141$ & $0.073$ \\
ECE $(\downarrow)$ & protein & $0.440$ & $0.649$ & $0.225$ & $0.567$ & $0.288$ & $0.325$ & $0.809$ & $0.565$ & $0.265$ \\
WS $(\downarrow)$ & protein & $0.592$ & $1.379$ & $0.156$ & $8.242$ & $0.182$ & $0.183$ & $1.483$ & $0.336$ & $0.245$ \\
\addlinespace[0.4em]
\hline
\addlinespace[0.4em]
RMSE $(\downarrow)$ & year & $0.800$ & $0.765$ & $0.789$ & $0.803$ & $0.775$ & $0.785$ & $0.785$ & $0.786$ & $0.812$ \\
NLL $(\downarrow)$ & year & $11.250$ & $12.064$ & $0.169$ & $1.861$ & $-0.023$ & $0.940$ & $7.952$ & $1.148$ & $0.477$ \\
ECE $(\downarrow)$ & year & $1.151$ & $1.003$ & $0.249$ & $0.230$ & $0.261$ & $1.043$ & $1.126$ & $1.219$ & $0.369$ \\
WS $(\downarrow)$ & year & $2.848$ & $2.543$ & $0.187$ & $0.163$ & $0.164$ & $0.539$ & $2.461$ & $0.625$ & $0.454$ \\
\addlinespace[0.5em]
\bottomrule
\end{tabular}
\end{adjustbox}
\label{tab:uci_eval_per_dataset}
\end{table*}


\subsection{Residual-uncertainty scatter plots}
\label{appendix:res_error}

\begin{figure}[htb]
    \centering
    \includegraphics[width=0.8\columnwidth]{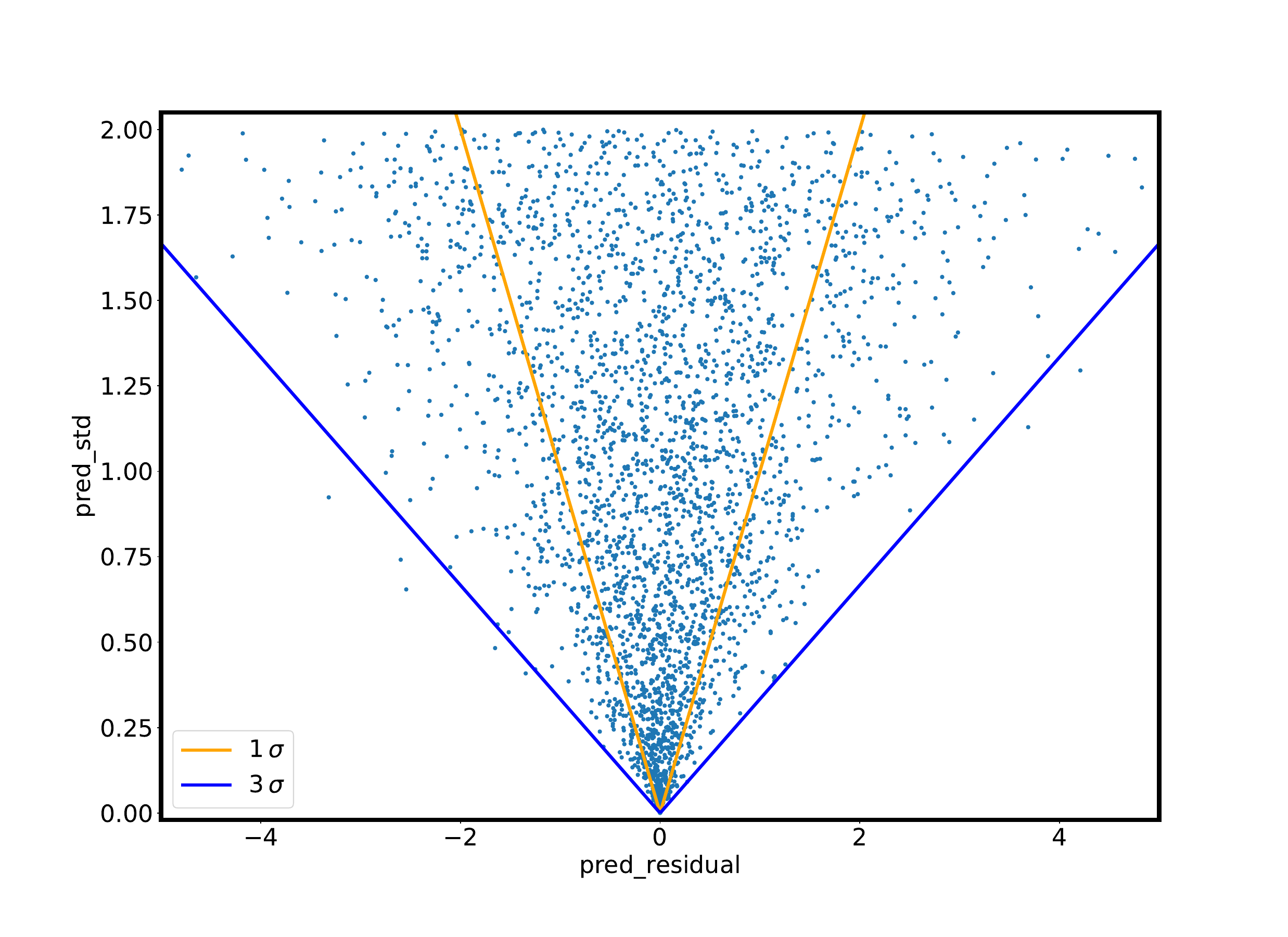} 
    \caption{Prediction residuals (x-axis) and predictive uncertainty (y-axis) for a \textit{hypothetical} ideal uncertainty mechanism. The Gaussian errors are matched by Gaussian uncertainty predictions at the exact same scale. $68.3\%$ of all uncertainty estimates (plot points) lie above the orange $1\sigma$-lines and $99.7 \%$ of them above the blue $3\sigma$-lines.}
    \label{fig:residual_error_gauss}
\end{figure}

\begin{figure*}[bth]
    \centering
    \includegraphics[width=1.0\textwidth]{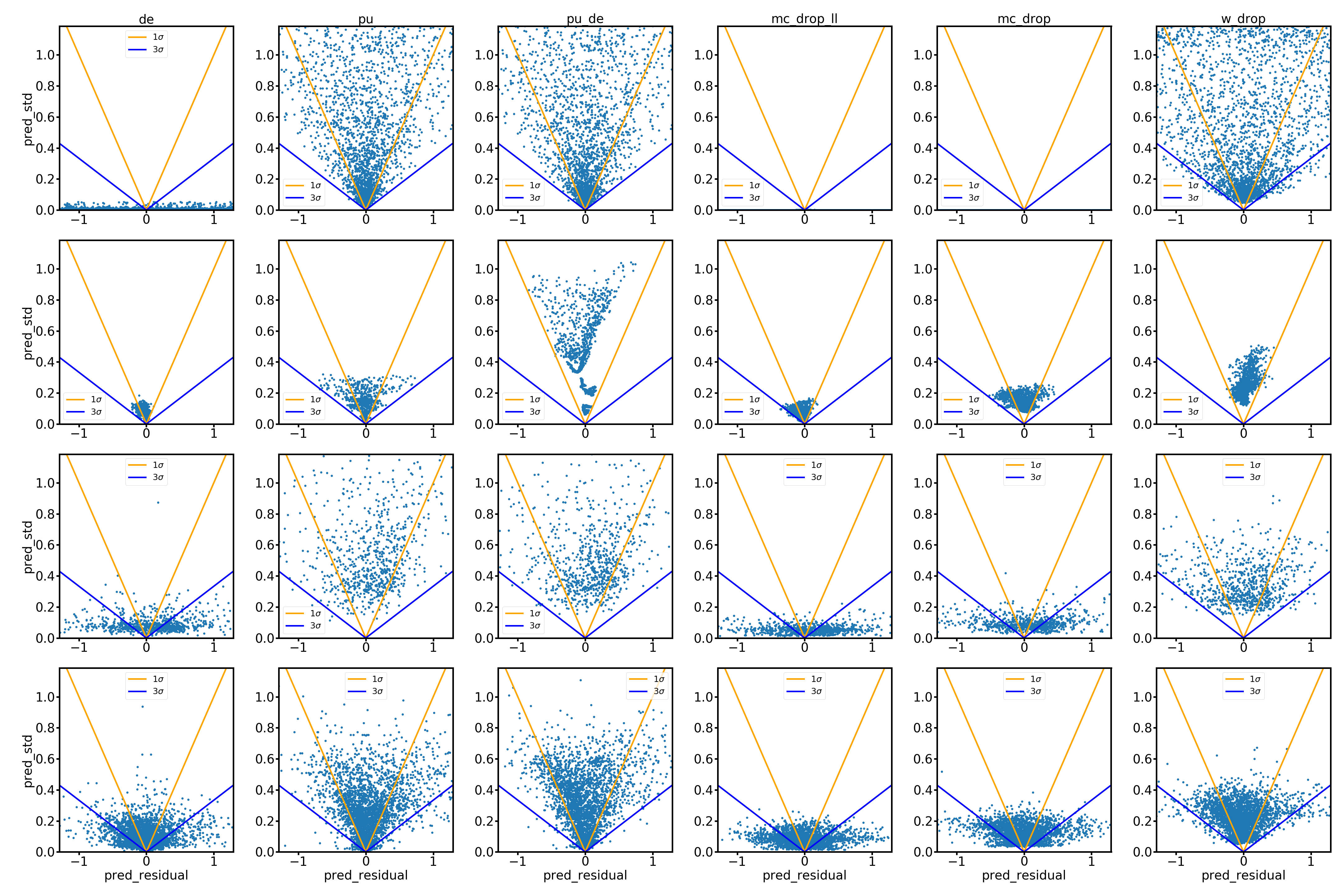}
    \caption{Prediction residuals (respective x-axis) and predictive uncertainty (respective y-axis) for different uncertainty mechanisms (columns) and datasets (rows). Each light blue dot in each plot corresponds to one test data point. Realistic uncertainty estimates should lie mostly above the blue $3\sigma$-lines. The datasets toy-noise, naval, abalone and superconduct are shown, from top to bottom.}
    \label{fig:appendix_uci_scatter}
\end{figure*}

Visual inspection of uncertainties can be helpful to understand their qualitative behavior.
We scatter model residuals $\mu_i - y_i$ (respective x-axis in Fig.\ \ref{fig:appendix_uci_scatter}) against model uncertainties $\sigma_i$ (resp. y-axis in Fig.\ \ref{fig:appendix_uci_scatter}).
For a \textit{hypothetical ideal} uncertainty mechanism, we expect $(y_i -\mu_i) \sim \mathcal{N}(0,\sigma_i)$, 
\ienows, model residuals following the predictive uncertainty distribution.
More concretely, $68.3\%$ of all $(y_i -\mu_i)$ would lie within the respective interval $[-\sigma_i,\sigma_i]$ and 99.7\% of all $(y_i -\mu_i)$ within $[-3\,\sigma_i, 3\,\sigma_i]$. Fig.\ \ref{fig:residual_error_gauss} visualizes this hypothetical ideal. It is generated as follows: We draw $3{,}000$ standard deviations $\sigma_i \sim \mathcal{U}(0,2)$ and sample residuals $r_i$ from the respective normal distributions, $r_i \sim \mathcal{N}(0,\sigma_i)$. The pairs $(r_i,\sigma_i)$ are visualized. By construction, uncertainty estimates now ideally match residuals in a distributional sense. 

Geometrically, the described Gaussian properties imply that $99.7\%$ of all scatter points, \egnows, in Fig.\ \ref{fig:appendix_uci_scatter}, should lie above the blue $3\sigma$ lines and $68.3\%$ of them above the yellow $1\sigma$ lines. 
For toy-noise, abalone and superconduct (first, third and fourth row in Fig.\ \ref{fig:appendix_uci_scatter}), PU, PU-DE and \ours qualitatively fulfill this requirement while MC, MC-LL and DE tend to underestimate uncertainties. This finding is in accordance with our systematic evaluation.
The naval dataset (second row in Fig.\ \ref{fig:appendix_uci_scatter}) poses an exception in this regard as all uncertainty methods lead to comparably convincing uncertainty estimates. The small test RMSEs of all methods on naval 
indicate relatively small aleatoric uncertainties and model residuals. Epistemic uncertainty might thus be a key driving factor and coherently MC, MC-LL and DE perform well.


\subsection{Object detection: systematic evaluation}
\label{appendix:squeeze}
\begin{figure}[htb]
    \centering
    \includegraphics[width=0.8\columnwidth]{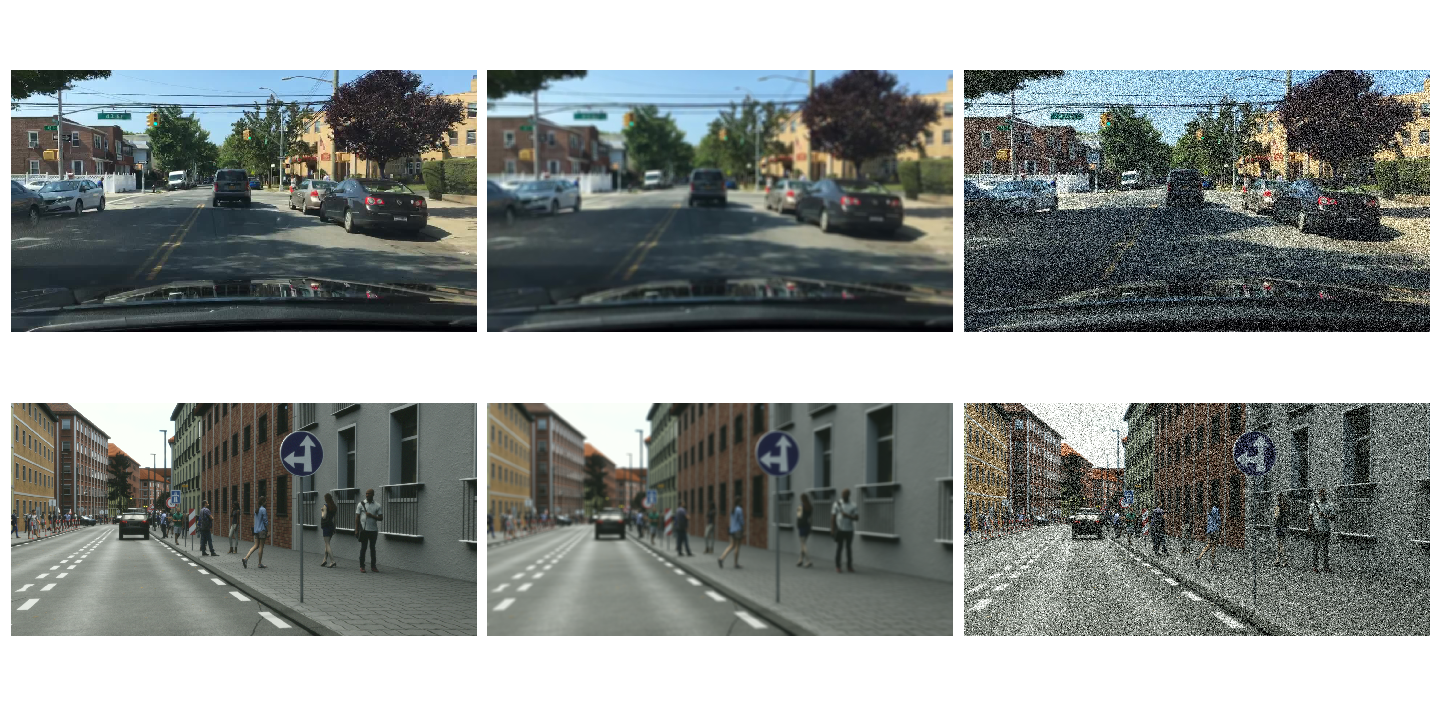} 
    \caption{Two exemplary object detection images from BDD100k (top row, real-world image) and SynScapes (bottom row, synthetic image), respectively. For each original image (left column), two corrupted versions are generated: a blurred one (middle column) and a noisy one (right column), see text for details.}
    \label{fig:od_images_blur_noise}
\end{figure}

We report basic information on the object detection (OD) datasets and their harmonization in the first paragraph of this subsection. Supplementary evaluations of SqueezeDet can be found subsequently in the second paragraph.
\paragraph{Details on OD datasets}
The six OD datasets we consider are diverse in multiple dimensions as they capture traffic scenes from three continents (Asia, Europe and North America) and cover a broad set of scenarios ranging from cities and metropolitan areas over country roads to highways (see Table \ref{tab:od_datasets_general_info}). They moreover differ in the average number of objects per image (see Table \ref{tab:od_dataset_stats}) that reaches its highest values for the simulation-based SynScapes dataset.\footnote{For A2D2, 2D bounding boxes are inferred from semantic segmentation ground truth.} Finally, both random and sequence-based train-test splits are considered. This variety is moreover reflected in the numerous object classes the different datasets provide. Their mappings to three main categories (`pedestrian', `cyclist', `vehicle') can be found in Table \ref{tab:od_class_mapping}. Rare or irregular classes are removed. For KITTI, we moreover discard `van', `truck' and `person-sitting', following the original SqueezeDet paper. To analyze uncertainty quality on distorted images, blurred and noisy versions of the test datasets are created. Fig. \ref{fig:od_images_blur_noise} shows these transformations for two exemplary images from BDD100k (top row) and SynScapes (bottom row), respectively.

\paragraph{Further results on SqueezeDet}
Coordinate-wise regression results and uncertainty scores for MC-SqueezeDet and W-SqueezeDet on KITTI are shown in Table \ref{tab:squeezedet_details}. While we observe noteworthy differences between coordinates, the relative ordering of MC-SqueezeDet and W-SqueezeDet for a given measure remains the same.

\begin{figure}[htb]
    \centering
    \includegraphics[trim=0 0 0 0, clip, width=0.7\columnwidth]{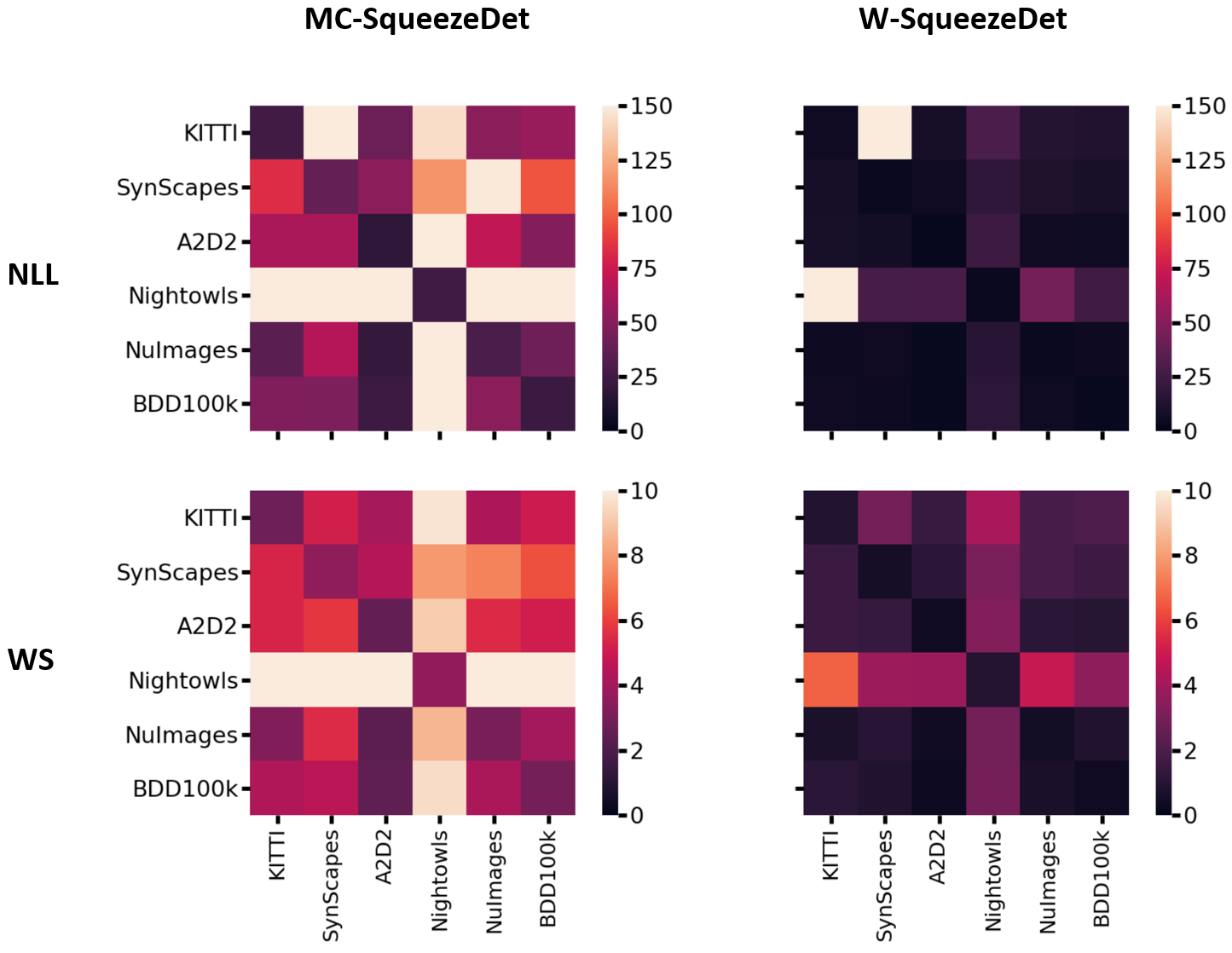} 
    \caption{In-data and out-of-data evaluation of MC-SqueezeDet (lhs) and W-SqueezeDet (rhs) on six OD datasets. We consider the negative log-likelihood (NLL, top row) and the Wasserstein measure (WS, bottom row). For each heatmap entry, the row label refers to the training dataset and the column label to the test dataset. Thus, diagonal matrix elements are in-data evaluations, non-diagonal elements are OOD analyses.}
    \label{fig:od_id-ood_nll_ws}
\end{figure}

Analyzing in-data and out-of-data NLL and WS values for all six datasets (see Fig.\ \ref{fig:od_id-ood_nll_ws}), we find results that qualitatively resemble those on ECE in Fig.\ \ref{fig:od_in-data_ood_mc_w}. W-SqueezeDet outperforms MC-SqueezeDet on the respective \iid test set and also under data shift. For both uncertainty approaches, some NLL values are affected by outliers.

\begin{figure}[htb]
    \centering
    \includegraphics[trim=130 20 160 20, clip, width=0.95\textwidth]{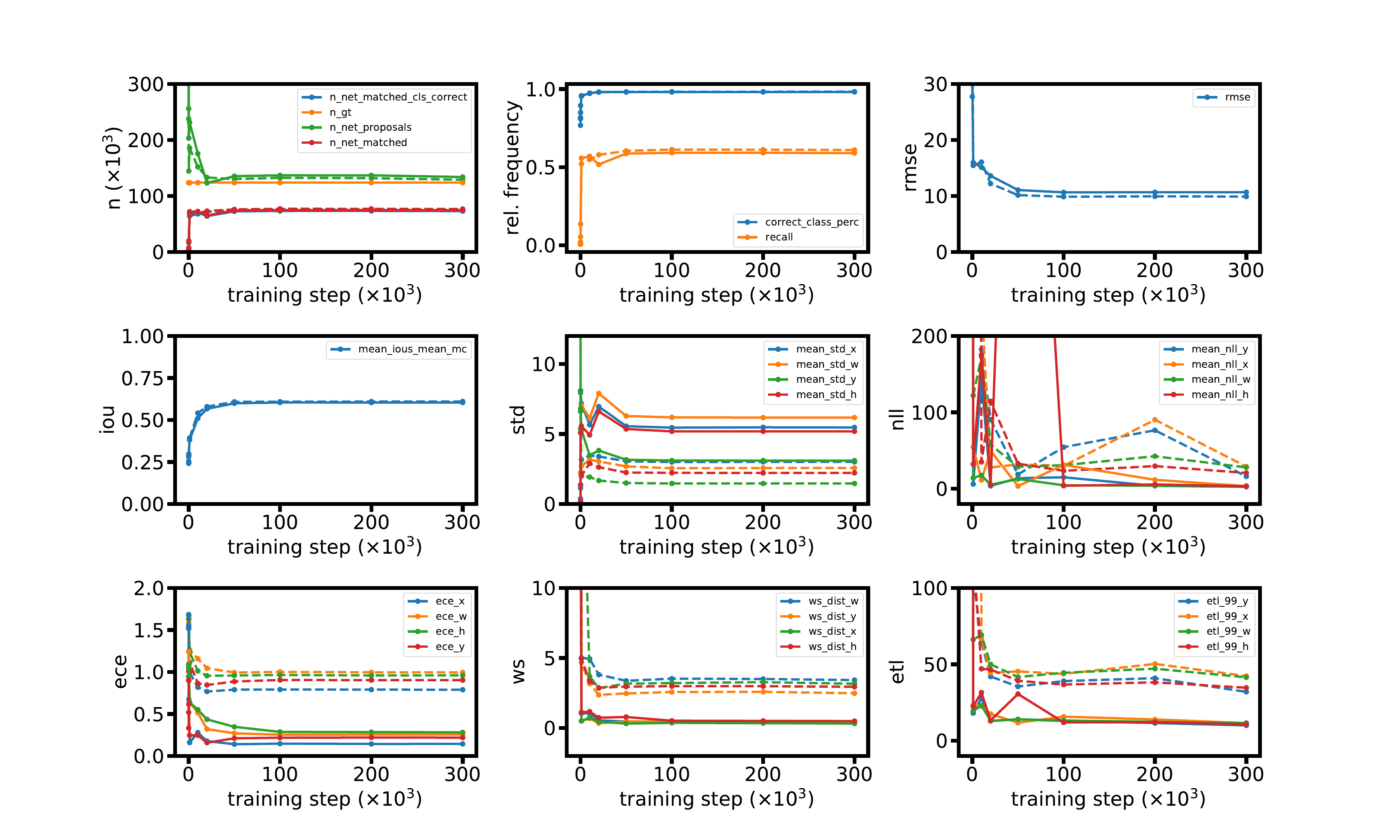} 
    \caption{Various test statistics of W-SqueezeDet (solid lines) and MC-SqueezeDet (dashed lines) during model optimization on the BDD100k dataset. We consider performance scores (recall, RMSE, IoU, see first and second row) and uncertainty measures (NLL, ECE, WS, ETL, see second and third row). W-SqueezeDet yields comparable task performance while providing clearly better uncertainty estimates.}
    \label{fig:od_stats_during_training_bdd}
\end{figure}

Finally, Fig.\ \ref{fig:od_stats_during_training_bdd} visualizes how various regression and uncertainty (test) scores evolve during model training on the BDD100k dataset. MC-SqueezeDet (dashed) and W-SqueezeDet (solid) `converge' with comparable speed (no changes to test RMSE and mIoU after 100{,}000 training steps) and reach similar final performances. W-SqueezeDet's explicit optimization of uncertainty estimates yields larger standard deviations (center panel) and smaller values for NLL, ECE, WS and ETL compared to MC-SqueezeDet (center right panel, bottom row). For the unbounded scores NLL, WS and ETL, W-SqueezeDet exhibits higher stability during training.

\begin{table}[th] 
\small
\centering
\caption{General information on the object detection datasets.}\label{tab:od_datasets_general_info}
\vspace{0.8em}
\setlength\tabcolsep{6pt} 
\begin{tabular}{l l l l l}
\toprule
  {dataset} & {place of data collection} & {type} & {train/test split} \\
\bottomrule
  \addlinespace[0.5em]
KITTI     &	metropolitan area of Karlsruhe & real & semi-custom (sequence-based) \\	
SynScapes &	simulation (only urban scenes) & synthetic & custom (random split) \\
A2D2	  & highways and cities in Germany & real & custom (sequence-based) \\
Nightowls &	several cities across Europe & real & pre-defined \\
NuImages  &	Boston and 3 diverse areas of Singapur & real & pre-defined	\\
BDD100k	  & New York, San Francisco Bay, Berkeley & real & pre-defined	\\
\bottomrule
  \addlinespace[0.8em]
\end{tabular}
\end{table}

\begin{table} 
\small
\centering
\caption{Harmonization of the object detection datasets. The various object classes of the six object detection datasets (rows) are grouped into the three main categories ``vehicle", ``pedestrian" and ``cyclist" (columns). Some classes are too rare or irregular and are thus discarded.}
\label{tab:od_class_mapping}
\vspace{0.8em}
\setlength\tabcolsep{6pt} 
\begin{tabular}{l p{4cm} p{2cm} p{2cm} p{4cm}}
\toprule
  {dataset} & {vehicle} & {cyclist} & {pedestrian} & {discarded} \\
\bottomrule
  \addlinespace[0.5em]
KITTI     & car & cyclist & pedestrian & van, truck, tram, person-sitting, misc, dontcare  \\	
SynScapes &	car, motorbike, truck, bus & cyclist & pedestrian & train  \\
A2D2	  & car, truck & cyclist & pedestrian & -  \\
Nightowls &	motorbike & cyclist & pedestrian & ignore-area \\
NuImages  & car, motorbike, truck, vehicle-other & cyclist & pedestrian & movable-object 	\\
BDD100k	  & car, motorbike, truck, bus, trailer, vehicle-other & cyclist & pedestrian, other-person & train, rider, traffic-light, traffic-sign	\\
\bottomrule
  \addlinespace[0.8em]
\end{tabular}
\end{table}

\begin{table*}[ht] 
\centering
\caption{Regression performance and uncertainty quality of SqueezeDet-type networks on KITTI train/test data. W-SqueezeDet is compared with MC-SqueezeDet.}
\label{tab:squeezedet_details}
\vspace{0.8em}
\begin{tabular}{l r r r r}
\toprule
{measure} & {MC-SqzDet} & {W-SqzDet} & {MC-SqzDet} & {W-SqzDet}\\\cmidrule(lr){2-3} \cmidrule(lr){4-5}
{~ } & \multicolumn{2}{c}{train} & \multicolumn{2}{c}{test} \\
\bottomrule
\addlinespace[0.5em]
mIoU $(\uparrow)$ & $\textbf{0.705}$ & $0.691$ & $\textbf{0.695}$ & $0.694$ \\
RMSE $(\downarrow)$ & $\textbf{8.769}$ & $9.832$ & $14.666$ & $\textbf{14.505}$ \\
\addlinespace[0.5em]
NLL$_x$ $(\downarrow)$ & $14.793$ & $\textbf{2.808}$ & $34.827$ & $\textbf{6.941}$ \\
NLL$_y$  & $6.135$ & $\textbf{2.170}$ & $13.364$ & $\textbf{3.808}$ \\
NLL$_w$  & $6.916$ & $\textbf{3.305}$ & $36.384$ & $\textbf{8.579}$ \\
NLL$_h$  & $6.146$ & $\textbf{2.796}$ & $18.241$ & $\textbf{5.908}$ \\
\addlinespace[0.5em]
ECE$_x$ $(\downarrow)$ & $0.560$ & $\textbf{0.148}$ & $0.748$ & $\textbf{0.330}$ \\
ECE$_y$ & $0.659$ & $\textbf{0.180}$ & $0.835$ & $\textbf{0.419}$ \\
ECE$_w$ & $0.523$ & $\textbf{0.147}$ & $0.888$ & $\textbf{0.520}$ \\
ECE$_h$  & $0.716$ & $\textbf{0.296}$ & $0.83$ & $\textbf{0.465}$ \\
\addlinespace[0.5em]                                              
WS$_x$ $(\downarrow)$ & $1.729$ & $\textbf{0.283}$ & $3.06$ & $\textbf{0.830}$ \\
WS$_y$ & $1.370$ & $\textbf{0.299}$ & $2.260$ & $\textbf{0.680}$ \\
WS$_w$  & $1.145$ & $\textbf{0.243}$ & $3.485$ & $\textbf{1.203}$ \\
WS$_h$  & $1.442$ & $\textbf{0.437}$ & $2.517$ & $\textbf{0.888}$ \\
\addlinespace[0.5em]   
ETL$_{0.99,x}$  & $34.443$ & $\textbf{9.316}$ & $55.772$ & $\textbf{21.310}$ \\
ETL$_{0.99,y}$  & $18.202$ & $\textbf{7.677}$ & $26.675$ & $\textbf{11.772}$ \\
ETL$_{0.99,w}$  & $19.914$ & $\textbf{10.835}$ & $53.408$ & $\textbf{23.202}$ \\
ETL$_{0.99,h}$  & $16.872$ & $\textbf{7.584}$ & $32.547$ & $\textbf{16.608}$ \\
\addlinespace[0.3em]
\bottomrule
\end{tabular}
\end{table*}

\section{Stability \wrt hyper-parameters \texorpdfstring{$p$}{p} and \texorpdfstring{$L$}{L}}
\label{appendix:hyper}

\ours possesses two hyper-parameters: the neuron drop rate $p$ and the sample size $L$ used to calculate the empirical estimates $\mu_{\bthetaTilde}(x_i)$ and $\sigma_{\bthetaTilde}(x_i)$. Here, we analyze the impact of these parameters on the quality of accordingly trained models.

\begin{figure*}[ht]
    \centering
    \includegraphics[trim=40 40 40 40, clip, width=1.0\textwidth]{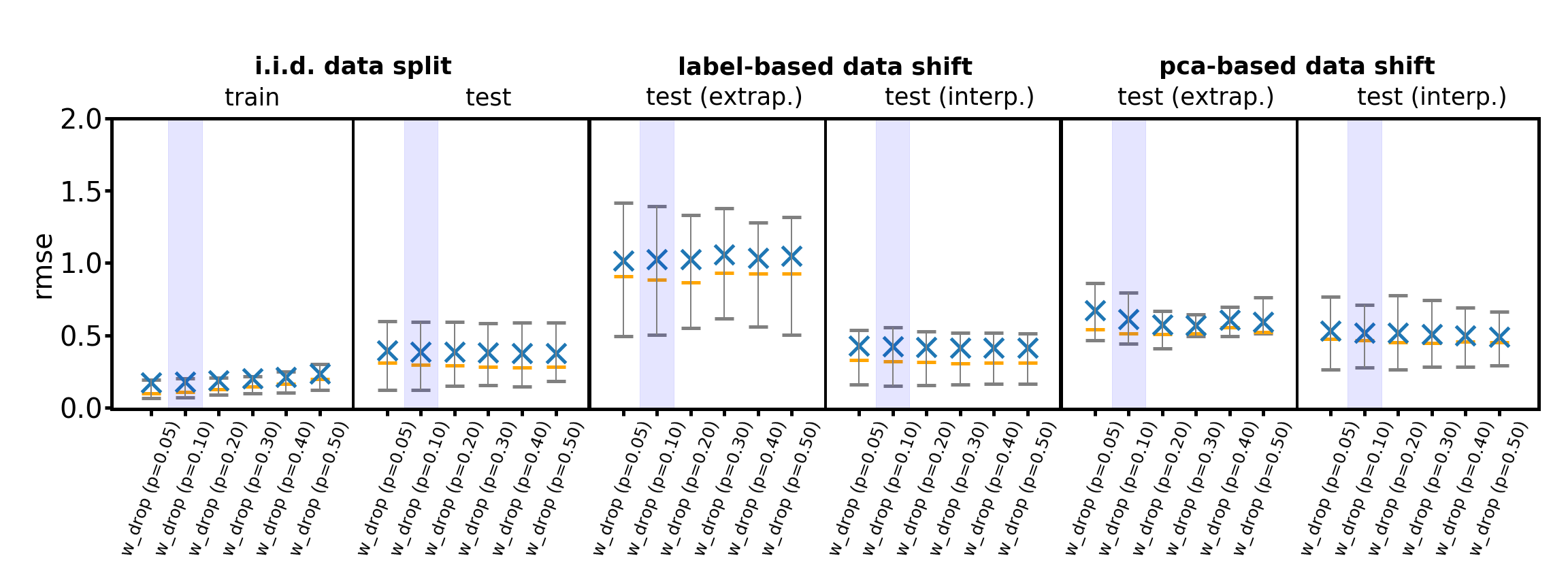}
    \includegraphics[trim=40 40 40 40, clip, width=1.0\textwidth]{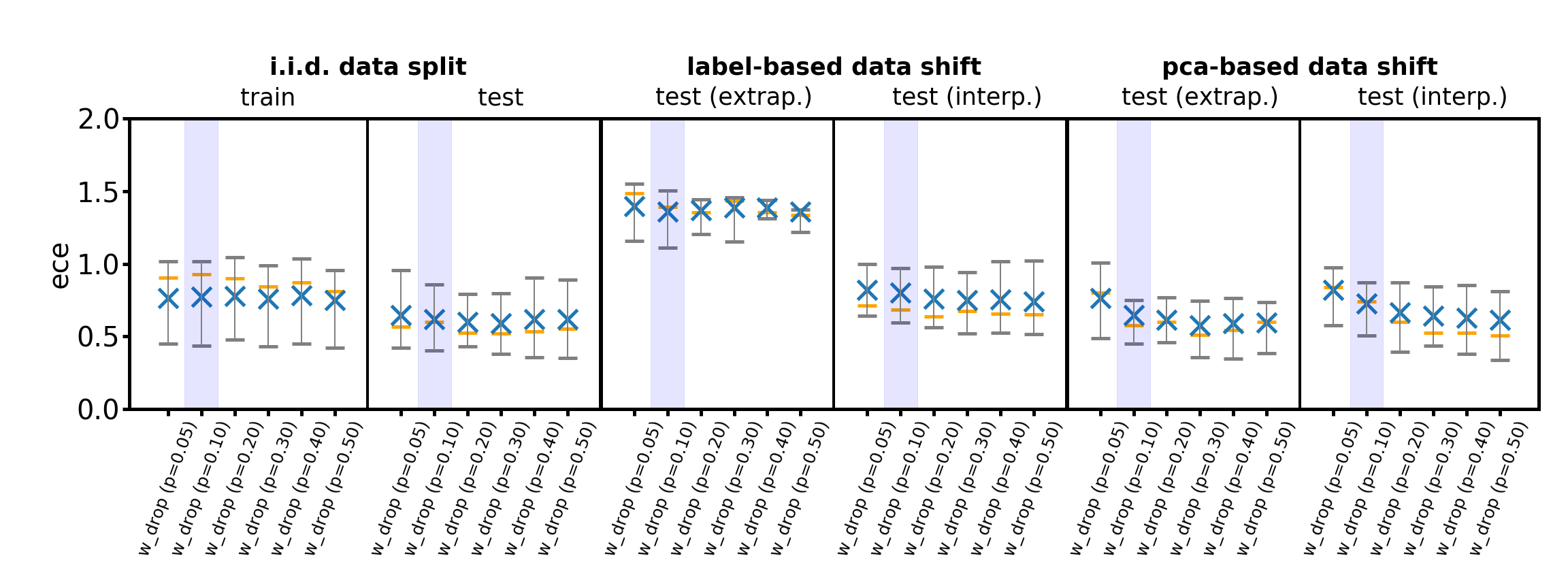}
    \caption{Dependence of Wasserstein dropout on drop rate $p$. Root-mean-square errors (RMSEs ($\downarrow$), top row) and expected calibration errors (ECEs ($\downarrow$), bottom row) are shown for neuron drop rates of $p = 0.05, 0.1, 0.2, 0.3, 0.4, 0.5$ under \iid conditions (first and second panel in each row) and under various kinds of data shift (third to sixth panel in each row, see text for details). W-dropout with $p=0.1$ (used for evaluations on toy and 1D regression data) is highlighted by a light blue background. Each blue cross is the mean over $10$ standard regression datasets. Orange line markers indicate median values. The gray vertical bars reach from the $25\%$ quantile (bottom horizontal line) to the $75\%$ quantile (top horizontal line).}
    \label{fig:hyper_p}
\end{figure*}

For $p = 0.05, 0.1, 0.2, 0.3, 0.4, 0.5$, we observe only relatively small differences in both RMSE (see top panel of Fig.\ \ref{fig:hyper_p}) and ECE (see bottom panel of Fig.\ \ref{fig:hyper_p}). On train data, RMSE slightly deteriorates with increasing $p$, \ienows, with decreasing complexity of the sub-networks. 
For ECE, we find minor improvements with growing drop rate which might be explained by the fact that the $L$ sub-networks in a given optimization step overlap less for higher $p$-values, thus allowing them to approximate the actual data distribution more closely. We choose $p = 0.1$ as the complexity of the resulting sub-networks is only mildly reduced compared to the deterministic full network.

\begin{figure*}[ht]
    \centering
    \includegraphics[trim=40 40 40 40, clip, width=1.0\textwidth]{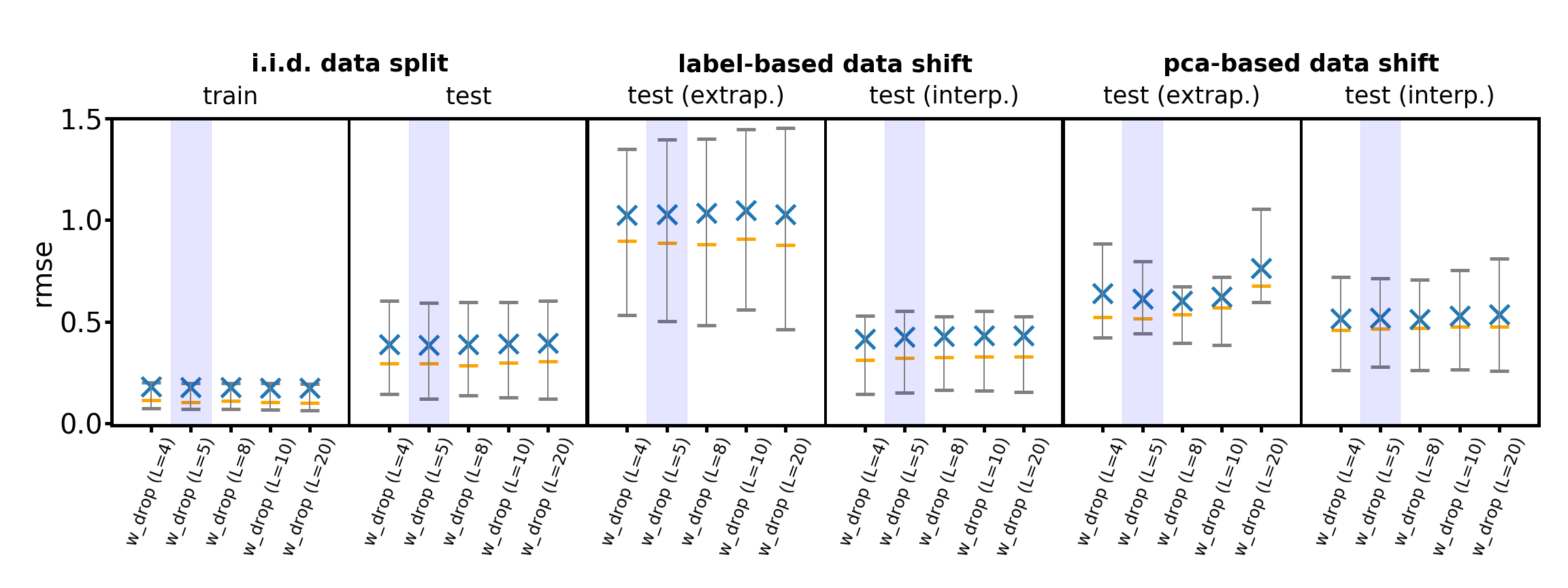}
    \includegraphics[trim=40 40 40 40, clip, width=1.0\textwidth]{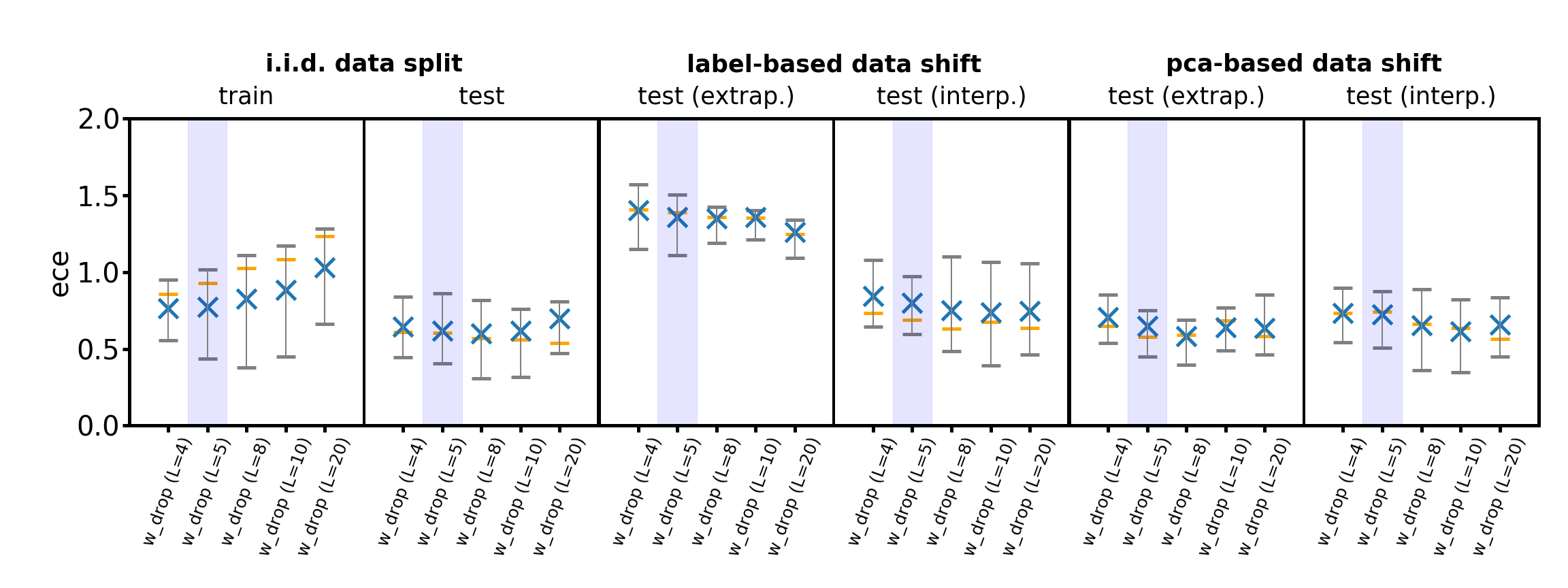}
    \caption{Dependence of Wasserstein dropout on sample size $L$. Root-mean-square errors (RMSEs ($\downarrow$), top row) and expected calibration errors (ECEs ($\downarrow$), bottom row) are shown for sample sizes of $L = 4, 5, 8, 10, 20$ under \iid conditions (first and second panel in each row) and under various kinds of data shift (third to sixth panel in each row, see text for details). W-dropout with $L=5$ (used throughout the rest of the paper) is highlighted by a light blue background. Each blue cross is the mean over $10$ standard regression datasets. Orange line markers indicate median values. The gray vertical bars reach from the $25\%$ quantile (bottom horizontal line) to the $75\%$ quantile (top horizontal line).}
    \label{fig:hyper_L}
\end{figure*}

Studying the impact of sample size $L = 4, 5, 8, 10, 20$, we find RMSE (see top panel of Fig.\ \ref{fig:hyper_L}) to be largely stable \wrt this parameter. For ECE (see bottom panel of \mbox{Fig.\ \ref{fig:hyper_L}}), train scores grow with $L$, indicating a certain over-estimation of the present aleatoric uncertainties. 
This artefact is not generalized to test data though, where we observe broadly similar mean values and 75\% quantiles. Under data shift, certain fluctuations of ECE occur as sample size $L$ changes, however there is no clear trend. We thus choose the rather small $L = 5$ to keep the computational overhead down.

\section{In-depth investigation of uncertainty measures}
\label{appendix:unc_measures}
In the following, we employ the Kolmogorov-Smirnov distance as a supplementary uncertainty score and compare it with expected calibration error (ECE) and Wasserstein distance (WS). Finally, limitations of negative log-likelihood (NLL) for uncertainty quantification are discussed.

\subsection{Dependencies between uncertainty measures}

\begin{figure*}[b]
    \centering
    \includegraphics[trim=0 0 0 0, clip, width=0.75\textwidth]{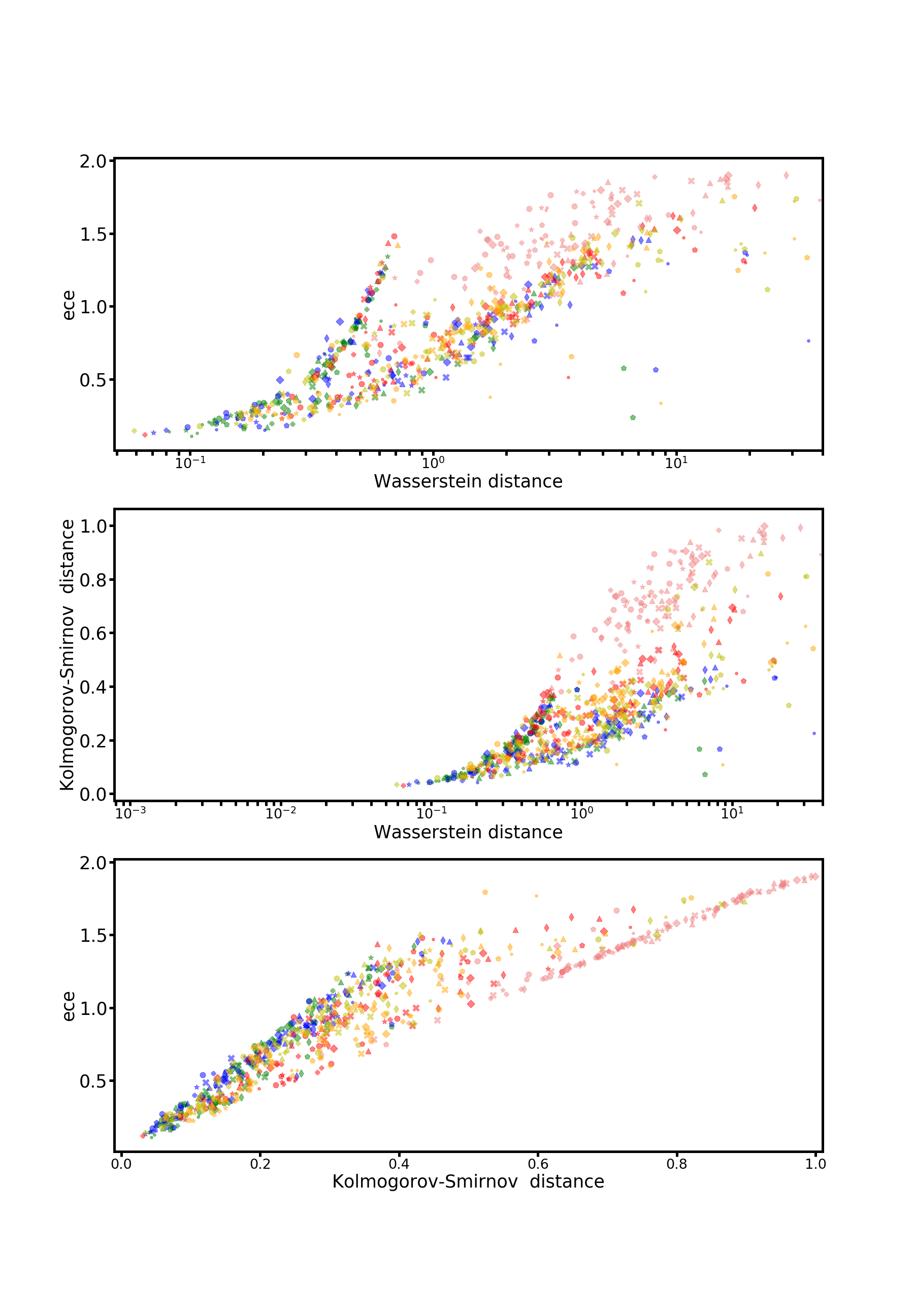} 
     \caption{Dependencies between the three uncertainty measures ECE, Wasserstein distance and Kolmogorov-Smirnov distance. Uncertainty methods are encoded via plot markers, data splits via color. Datasets are not encoded and cannot be distinguished (see text for more details). Each plot point corresponds to a cross-validated trained network. The clearly visible deviations from ideal correlations point at the potential of these uncertainty measures to complement one another.} 
    \label{fig:corrs_uncertainty_measures}
\end{figure*}

Extending the analysis of empirically observed dependencies between WS and ECE in \mbox{Fig.\ \ref{fig:ece_ws_uci_data}}, we additionally consider Kolmogorov-Smirnov (KS) distances \citep{stephens1974edf} 
in \mbox{Fig.\ \ref{fig:corrs_uncertainty_measures}} (middle and bottom panel). 
These KS-distances are calculated between samples of normalized residuals and a standard Gaussian.
Different from the Wasserstein distance, the KS-distance is not transport-based but determined by the largest distance between the empirical CDFs of the two samples.
It is therefore bounded to $[0,1]$ and unable to resolve differences between two samples that both strongly deviate from a standard Gaussian. Again, we find the dependencies between these measures to clearly deviate from ideal correlation.

The data splits in Fig.\ \ref{fig:ece_ws_uci_data} and Fig.\ \ref{fig:corrs_uncertainty_measures} are color-coded as follows: train is green, test is blue, PCA-interpolate is green-yellow, PCA-extrapolate is orange-yellow, label-interpolate is red and label-extrapolate is light red. The mapping between uncertainty methods and plot markers reads: SWAG is `triangle', MC is `diamond', MC-LL is `thin diamond', DE is `cross', PU is `point', PU-DE is `star', PU-MC is `circle', PU-EV is `pentagon' and \ours is `plus'. The data base of this visualization are the 14 standard regression datasets. Some Wasserstein distances lie above the x-axis cut-off and are thus not visualized.


\subsection{Discussion of NLL as a measure of uncertainty}
\label{appendix:nll}

Typically, DNNs using uncertainty are often evaluated in terms of their negative log-likelihood (NLL).
This property is affected not only by the uncertainty, but also by the DNNs performance.
Additionally, it is difficult to interpret, sometimes leading to counterintuitive results, which we want to elaborate on here.
As a first example, take the likelihood of two datasets $x_1=\{0\}$ and $x_2=\{0.5\}$, each consisting of a single point, with respect to a normal distribution $\mathcal{N}(0,1)$.
Naturally, we find $x_1$ to be located at the maximum of the considered normal distribution and deem it the more likely candidate.
But, if we extend these datasets to more than single points, \ienows, $\tilde x_1= \{0,0.1,0,-0.1,0\}$ and $\tilde x_2=\{0.5,-0.4,0,-1.9,-0.7\}$, it becomes obvious that $\tilde x_2$ is much more likely to follow the intended Gaussian distribution.
Nonetheless, $\text{NLL}(\tilde x_2)\approx 1.4 > 0.9 \approx \text{NLL}(\tilde x_1)$, where
\begin{equation}
    \text{NLL}(y):=\log{\sqrt{2\pi\sigma^2}}+\frac{1}{N}\sum_{i=1}^N \frac{(y_i-\mu)^2}{2\sigma^2}\,.
    \label{eq:appxNLL}
\end{equation}
This may be seen as a direct consequence of the point-wise definition of NLL, which does not consider the distribution of the elements in $\tilde x_i$.
From this observation also follows that a model with high prediction accuracy will have a lower NLL score as a worse performing one if uncertainties are predicted in the same way.
Independent of whether those reflected the ``true'' uncertainty in either case.
This issue can be further substantiated on a second example.
Consider two other datasets $z_1,z_2 $ drawn \iid from Gaussian distributions $\mathcal{N}(0,\sigma_i)$ with two differing values $\sigma_1\!<\!\sigma_2$.
If we determine the NLL of each with respect to its own distribution the offset term in equation (\ref{eq:appxNLL}) leads to $\text{NLL}(z_2)=\text{NLL}(z_1)+\log{(\sigma_2/\sigma_1)}$ with $\log{(\sigma_2/\sigma_1)}>0$.
Although both accurately reflect their own distributions, or uncertainties so to speak, the narrower $z_1$ is more ``likely''.
This offset makes it difficult to assess reported NLL values for systems with heteroscedastic uncertainty.
While smaller is typically ``better'', it is highly data- (and prediction-)dependent which value is good in the sense of a reasonable correlation between performance and uncertainty.

\end{document}